\documentclass[final]{cvpr}

\usepackage{times}
\usepackage{epsfig}
\usepackage{graphicx}
\usepackage{amsmath}
\usepackage{amssymb}
\usepackage{verbatim}
\usepackage{bm}
\usepackage{color}
\usepackage{subfig} 
\usepackage{caption}
\usepackage{booktabs}
\usepackage{subfig}
\usepackage[pagebackref=true,breaklinks=true,colorlinks,bookmarks=false]{hyperref}
\usepackage[ruled,vlined,commentsnumbered]{algorithm2e}

\def\cvprPaperID{1576} % *** Enter the CVPR Paper ID here

\def\confYear{CVPR 2021}
\def\comp{\ensuremath\mathop{\scalebox{.6}{$\circ$}}}
\newcommand\gm{{GMTracker}}

\newcommand{\MCG}{\mathcal{G}}
\newcommand{\MCV}{\mathcal{V}}
\newcommand{\MCE}{\mathcal{E}}
\newcommand{\MPI}{\mathbf{\Pi}}
\newcommand{\MA}{\mathbf{A}}

\newcommand{\MBR}{\mathbb{R}}
\newcommand{\MB}{\mathbf{B}}
\newcommand{\Mh}{\mathbf{h}}
\newcommand{\Mm}{\mathbf{m}}
\newcommand{\MCT}{\mathcal{T}}
\newcommand{\MCD}{\mathcal{D}}
\newcommand{\MCA}{\mathcal{A}}
\newcommand{\MCF}{\mathcal{F}}
\DeclareMathOperator*{\minimize}{minimize}
\DeclareMathOperator*{\diag}{diag}
\DeclareMathOperator*{\subjectto}{subject\;to}
\newcommand{\norm}[1]{\| #1 \|}

\begin{document}
\pagestyle{empty}
%%%%%%%%% TITLE
\def\paperTitle{Learnable Graph Matching: Incorporating Graph Partitioning with Deep Feature Learning for Multiple Object Tracking }
% \title{Learnable Graph Matching: Incorporating Graph Partitioning with Deep Feature Learning for Multiple Object Tracking}
\title{\paperTitle}
\author{
        Jiawei He$^{1,3}$ \quad 
        Zehao Huang$^{2}$ \quad 
        Naiyan Wang$^{2}$ \quad 
        Zhaoxiang Zhang$^{1,3,4}$ \\
        $^{1}$ Institute of Automation, Chinese Academy of Sciences (CASIA)\quad
        $^{2}$ TuSimple\\
        $^{3}$ School of Artificial Intelligence, University of Chinese Academy of Sciences (UCAS)\\
        $^{4}$ Centre for Artificial Intelligence and Robotics, HKISI\_CAS\\
        {\tt\small
 \{hejiawei2019, zhaoxiang.zhang\}@ia.ac.cn \{zehaohuang18, winsty\}@gmail.com}
        }

\maketitle
\thispagestyle{empty}

%%%%%%%%% ABSTRACT
\begin{abstract}
Data association across frames is at the core of Multiple Object Tracking (MOT) task. This problem is usually solved by a traditional graph-based optimization or directly learned via deep learning. 
Despite their popularity, we find some points worth studying in current paradigm:
1) Existing methods mostly ignore the context information among tracklets and intra-frame detections, which makes the tracker hard to survive in challenging cases like severe occlusion. 2) The end-to-end association methods solely rely on the data fitting power of deep neural networks, while they hardly utilize the advantage of optimization-based assignment methods. 3) The graph-based optimization methods mostly utilize a separate neural network to extract features, which brings the inconsistency between training and inference.
Therefore, in this paper we propose a novel learnable graph matching method to address these issues.
Briefly speaking, we model the relationships between tracklets and the intra-frame detections as a general undirected graph. Then the association problem turns into a general graph matching between tracklet graph and detection graph.
Furthermore, to make the optimization end-to-end differentiable, we relax the original graph matching into continuous quadratic programming and then incorporate the training of it into a deep graph network with the help of the implicit function theorem. Lastly, our method {\gm}, achieves state-of-the-art performance on several standard MOT datasets. Our code will be available at \url{https://github.com/jiaweihe1996/GMTracker}.
\end{abstract}

%%%%%%%%% BODY TEXT

\vspace{-10pt}
\section{Introduction}
% \begin{comment}
\begin{figure}
    \centering
        \includegraphics[width=\linewidth]{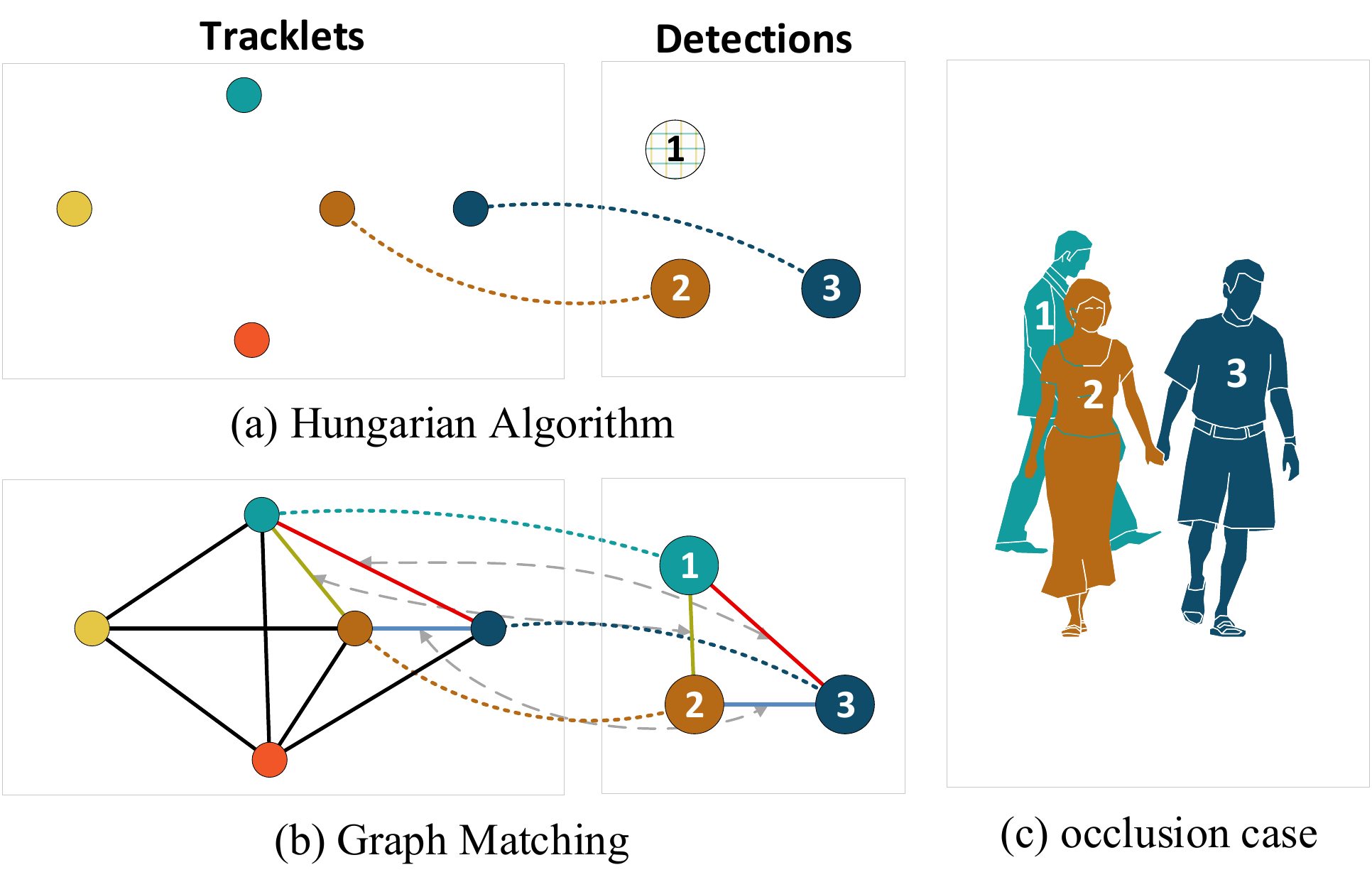}
    \caption{An illustration of intra-graph relationship used in our graph matching formulation. We utilize the second-order edge-to-edge similarity to model the group activity, which is more robust under heavy occlusion. Note that not all vertices in tracklet graph can be matched to the detection graph because they disappear in the current frame.}
    \vspace{-10pt}
    \label{fig1}
\end{figure}
% \end{comment}
Multiple Object Tracking (MOT) is a fundamental task that aims at associating the same object across successive frames in a video clip. A robust and accurate MOT algorithm is indispensable in broad applications, such as autonomous driving and video surveillance.
The \textit{tracking-by-detection} is currently the dominant paradigm in MOT. This paradigm consists of two steps: (1) obtaining the bounding boxes of objects by detection frame by frame; (2) generating trajectories by associating the same objects between frames. With the rapid development of deep learning based object detectors, the first step is largely solved by the powerful detectors such as~\cite{lin2017feature, lin2017focal}. 
As for the second one, recent MOT works focus on improving the performance of data association mainly from the following two aspects: (1) formulating the association problem as a combinatorial graph partitioning problem and solve it by advanced optimization techniques \cite{berclaz2011multiple, bewley2016simple, wang2017learning, braso2020learning, xu2020train, hornakova2020lifted}; (2) improving the appearance models by the power of deep learning\cite{kuo2011does, yang2012online, leal2016learning, wojke2017simple}. Although very recently, there are works~\cite{zhu2018online, sun2019deep, braso2020learning, xu2020train} trying to unify feature learning and data association into an end-to-end trained neural network, these two directions are almost isolated so that these recent attempts hardly utilize the progress from the combinatorial graph partitioning.

In the graph view of MOT, each vertex represents a detection bounding box or a tracklet, while the edges are constructed between the vertices across different frames to represent the similarities between them. Then the association problem can be formulated as a min-cost flow problem \cite{zhang2008global}. %The cost assigned to an edge represents the distance between objects, which can be defined as the similarity of appearance features of two objects \cite{wojke2017simple} or the intersection-over-union (IOU) between two bounding-boxes \cite{bewley2016simple}. 
The most popular used method is to construct a bipartite graph between two frames and adopt Hungarian algorithm \cite{kuhn1955hungarian} to solve it. This online strategy is widely used in practice because of its simplicity \cite{bewley2016simple, wojke2017simple}. 
Nevertheless, these methods are not robust to occlusions due to the lack of history trajectories and long term memory. 
This problem is traditionally solved by constructing graph from multiple frames, and then deriving the best association based on the optimal solution of this min-cost flow problem~\cite{zhang2008global}.
%Recently, MPNTrack \cite{braso2020learning} proposed to partition the graph into trajectories by supervised learning directly. %Message Passing Network (MPN) \cite{gilmer2017neural} is adopted to aggregate features across the graph. 
%With the help of supervised learning, the proposed method yields significant improvement on MOT benchmarks \cite{leal2015motchallenge, milan2016mot16}. %However, it is an offline method and does not utilize the relationship between trajectories since there are not connections among nodes in the same frame in the whole graph.

All these existing works focus on finding the best matching across frames, but ignoring the context within the frame.
In this paper, we argue that the relationship between the vertices within the same frame is also crucial for some challenging cases in MOT. 
For example, we can match an occluded object to the correct tracklet solely by the past relationships with neighborhood objects. Fig.~\ref{fig1} just shows such an example.
Interestingly, these pairwise relationships within the same frame can be represented as edges in a general graph.
To this end, the popular bipartite matching across frames can be updated to general graph matching between them. 
To further integrate this novel assignment formulation with powerful feature learning, we first relax the original formulation of graph matching \cite{LawlerMS63, kbqap} to a quadratic programming, and then derive a differentiable QP layer based on the KKT conditions and the implicit function theorem for the graph matching problem, inspired by the OptNet \cite{amos2017optnet}. Finally, the assignment problem can be learned in synergy with the features. %Not surprisingly, our \gm\ achieves state-of-the-art performance on the publicly available MOT datasets. 

% these informations between nodes in the same frame are important for data association.  Based on this motivation, we propose a novel online tracker named \gm. In our method, we construct two graphs, tracklet graph which stores the information of history trajectories and detection graph which is constructed by objects on current frame. Then a message passing network is adopted cross these two graph to enhance the features of nodes and edges. Finally, we do graph matching between these two graphs to find the assignment of detections to trajectories. To make the whole framework end-to-end learnable, we relax the original graph matching's QAP formulation \cite{LawlerMS63, LoiolaEJOR07} to a quadratic programming, and build the graph matching layer as a QP layer \cite{amos2017optnet}. 
%The whole framework contains three main parts: 
%1) tracklet graph and detection graph construction; 
%2) feature embedding and message passing between two graphs; 
%3) graph matching and rounding. 

Overall, our work has the following contributions:

\begin{itemize}
    \item Instead of only focusing on the association across frames, we emphasize the importance of intra-frame relationships. Particularly, we propose to represent the relationships as a general graph, and formulate the association problem as general graph matching.
    \item To solve this challenging assignment problem, and further incorporate it with deep feature learning, we derive a differentiable quadratic programming layer based on the continuous relaxation of the problem, and utilize implicit function theorem and KKT conditions to derive the gradient w.r.t the input features during back-propagation.
    \item We evaluate our proposed {\gm} on the large scale open benchmark. Our method could remarkably advance the state-of-the-art performance in terms of association metric such as IDF1.
\end{itemize}

\section{Related Work}
%Multiple object tracking task aims to find trajectories of different objects in a video clip. In most methods of the MOT task, a tracking-by-detection paradigm is applied, which means they should detect objects frame by frame and then associate the detection results using matching algorithms. Most mainstream methods focus on getting accurate detection results, extracting robust features from detected bounding boxes, and matching the bounding boxes belonging to the same track.

\noindent{\bf Data association in MOT.}
The data association step in \textit{tracking-by-detection} paradigm is generally solved by probabilistic filter or combinatorial optimization techniques.
Classical probabilistic approach includes JPDA \cite{bar1990tracking} and MHT \cite{reid1979algorithm}. The advantage of this approach is to keep all the possible candidates for association, and remain the chance to recover from failures. Nevertheless, their costs are prohibitive if no approximation is applied \cite{hamid2015joint, kim2015multiple}. 
For combinatorial optimization, traditional approach include bipartite matching \cite{bewley2016simple}, dynamic programming \cite{fleuret2007multicamera}, min-cost flow \cite{zhang2008global, berclaz2011multiple} and conditional random field \cite{yang2011learning}. Follow-up works tried to adopt more complex optimization methods \cite{zamir2012gmcp, tang2015subgraph}, reduce the computational cost \cite{pirsiavash2011globally, tang2016multi} or promote an online setting from them \cite{choi2015near, wang2016tracklet}.
%In the pre-deep learning era, several works \cite{zhang2008global, pirsiavash2011globally} propose a global data association approach by mapping the association problem as a cost-flow network with a non-overlap constraint on trajectories. \cite{zamir2012gmcp} and \cite{tang2015subgraph} adopt more complex optimization such as minimum cliques and subgraph multicut. SORT \cite{bewley2016simple} proposes a simple online framework for realtime MOT. Hungarian matching algorithm \cite{kuhn1955hungarian} is used for frame-by-frame data association.

\noindent{\bf Deep learning in MOT.}
Early works of deep learning in MOT such as \cite{wojke2017simple,leal2016learning, sadeghian2017tracking,Li_2020_WACV} mostly focus on learning a better appearance model for each object. Then by the advance of object detection and multi-task learning, several works \cite{bergmann2019tracking, lu2020retinatrack, zhou2020tracking, zhang2020fair} combine detection and tracking in the same framework. More recently, several works tried to bridge the graph optimization and end-to-end deep learning \cite{jiang2019graph, braso2020learning, xu2020train, Li_2020_WACV, hornakova2020lifted}. \cite{jiang2019graph} adopts Graph Neural Network (GNN) to learn an affinity matrix in a data-driven way. MPNTrack \cite{braso2020learning} introduces a message passing network to learn high-order information between vertices from different frames. \cite{Li_2020_WACV} constructs two graph networks to model appearance and motion features, respectively. 
LifT \cite{hornakova2020lifted} proposes a lifted disjoint path formulation for MOT, which introduces lifted edges to capture long term temporal interactions.
%Above methods mainly focus on modeling the connection of nodes between different frames, while ignore the graph context with in a frame. To address this issue, we adopt graph matching to find the data association. Through combining the edges information between nodes in the same frame, we can generate more robust matching results.
 
\noindent{\bf Neighborhood and context information in MOT.} Pedestrians usually walk in a group, so the motion of them are highly clustered. Modeling neighborhood and context relationships may provide important clues for the MOT task. Several early works \cite{rodriguez2009tracking,hu2008detecting} consider the group model as a prior in motion model for the crowd scenes. 
\cite{yoon2016online} considers the distance between detections as the neighborhood relationship during the data association.
% In the proposed group model, a group should have a similar motion, stay close with neighbors and keeping a distance from others. 
However, these hand-crafted priors can be quite limited in complicated scenes. Recently, many methods~\cite{liang2020enhancing,peng2020tpm,liugsm,lan2016online} learn appearance and geometric information by passing messages among neighbors. However, their goal is still to enhance the appearance features. They do not consider them explicitly in the association across frames.
In this work, we propose to explicitly consider the neighborhood context in data association via differentiable graph matching and use it to guide the feature learning in an end-to-end manner. 
% In other tasks, such as person search~\cite{yan2019learning} and trajectory prediction~\cite{xu2018encoding}, the collaboration between neighbors or in a group is also a common consideration. 
%However, for the MOT task, combining the graph model and neighbor information, despite the limited application, graph matching, which considers edge similarity, is a kind of better representation. 

\noindent{\bf Graph matching and Combinatorial Optimization.}
%Combinatorial optimization is the optimization problem whose optimization variable is discrete, such as Traveling Salesman Problem, Knapsack Problem, etc. The graph matching problem is also a combinatorial optimization problem. 
Pairwise graph matching, or more generally Quadratic Assignment Problem (QAP), has wide applications in various computer vision tasks \cite{vento2013graph}.
Compared with the linear assignment problem that only considers vertex-to-vertex relationship, pairwise graph matching also considers the second-order edge-to-edge relationship in graphs. 
The second-order relationship makes matching more robust. However, as shown in \cite{hartmanis1982computers}, this problem is an NP-hard problem. 
There is no polynomial solver like Hungarian algorithm~\cite{kuhn1955hungarian} for the linear assignment problem. In the past decades, many works focus on making the problem tractable by relaxing the original QAP problem~\cite{leordeanu2005spectral,schellewald2005probabilistic,torr2003solving}. Lagrangian decomposition~\cite{swoboda2017study} and factorized graph matching~\cite{facgm} are two representative ones. 

In MOT task, the application of graph matching is very limited. To the best of our knowledge, \cite{hu2020dual} is the first to formulate the MOT task as a graph matching problem and use dual L1-normalized tensor power iteration method to solve it. Different from \cite{hu2020dual} that directly extracts the features from an off-the-shelf neural network, we propose to guide the feature learning by the optimization problem, which can both enjoy the power of deep feature learning and combinatorial optimization. This joint training manner of representation and optimization problem also eliminate the inconsistencies between the training and inference.
%Algorithm complexity will be extremely high when the number of nodes in the graph increases. 

% Recently, an emerging trend is to incorporate graph matching and other combinatorial optimization problems into deep learning. 
To incorporate graph matching into deep learning, one stream of work is to treat the assignment problem as a supervised learning problem directly, and use the data fitting power of deep learning to learn the projection from input graphs to output assignment directly ~\cite{wang2019learning,Yu2020Learning}.
Another more theoretically rigorous is to relax the problem to a convex optimization problem first, and then utilize the KKT condition and implicit function theorem to derive the gradients w.r.t all variables at the optimal solution ~\cite{barratt2018differentiability}. As shown in ~\cite{amos2017optnet}, the universality and transferability of the latter approach are much better than the first one. 
Thus, in this paper, we derive a graph matching layer based on this spirit to solve the challenging graph matching problem in MOT.
% GMN~\cite{zanfir2018deep} is the first differential and end-to-end learned model to solve the graph matching problem. It makes the parameters in feature extractor, affinity matrix end-to-end learnable with a matching based loss function. Many works follow and improve it by aggregating features using graph neural network~\cite{wang2019learning} and modifying the loss function~\cite{Yu2020Learning}. 

\section{Graph Matching Formulation for MOT}
\label{sec:relax}
In this section, we will formulate the multiple object tracking problem as a graph matching problem. Instead of solving the original Quadratic Assignment Problem (QAP), we relax the graph matching formulation as a convex quadratic programming (QP) and extend the formulation from the edge weights to the edge features. The relaxation facilitates the differentiable and joint learning of feature representation and combinatorial optimization.
\begin{comment}
\textcolor{red}{In this section, we will describe the formulation of graph matching and how to relax it as a convex quadratic programming (QP) problem. Then we will show how to extend the formulation from edge weights to edge features.}
\end{comment}
\subsection{Detection and Tracklet Graphs Construction}
\label{sec:construct}
As an online tracker, we track objects frame by frame. 
In frame $t$, we define $\MCD^t=\{D_1^t, D_2^t,\cdots, D_{n_d}^t\}$ as the set of detections in current frame and $\MCT^t=\{T_1^t, T_2^t, \cdots, T_{n_t}^t\}$ as the set of tracklets obtained from past frames. $n_d$ and $n_t$ denote the number of detected objects and tracklet candidates. A detection is represented by a triple $D_p^t=(\mathbf{I}_p^t, \mathbf{g}_p^t, t)$, where $\mathbf{I}_p^t$ contains the image pixels in the detected area, $\mathbf{g}_p^t=(x_p^t,y_p^t,w_p^t,h_p^t)$ is a geometric vector including the central location and size of the detection bounding box. Each tracklet contains a series of detected objects with the same tracklet id. With a bit abuse of notations, the generation of $T_{id}^t$ can be represented as $T_{id}^{t} \gets T_{id}^{t-1} \cup \{D^{t-1}_{(id)}\}$, which means we add $D^{t-1}_{(id)}$ to the tracklet $T_{id}^{t-1}$.

Then we define the detection graph in frame $t$ as $\MCG_D^t=(\MCV_D^t, \MCE_D^t)$ and the tracklet graph up to the frame $t$ as $\MCG_T^t=(\MCV_T^t, \MCE_T^t)$. Each vertex $i \in \MCV_D^t$ and vertex $j\in \MCV_T^t$ represents the detection $D_i^t$ and the tracklet $T_j^t$, respectively. The $e_u=(i,i')$ is the edge in $\MCE_D^t$ and $e_v=(j,j')$ is the edge in $\MCE_T^t$. Both of these two graphs are complete graphs. Then the data association in frame $t$ can be formulated as a graph matching problem between $\MCG_D^t$ and $\MCG_T^t$. For simplicity, we will ignore $t$ in the following sections.

%In general, the \textbf{intuition} of our derivation below is that we relax the basic formulation of graph matching to a convex quadratic programming (QP), and it can become a QP layer in our learnable pipeline. To expand the formulation from edge weights to edge features, we finally have a formulation Eq.\ref{finalQP}.
\subsection{Basic Formulation of Graph Matching}
Given the detection graph $\MCG_D$ and the tracklet graph $\MCG_T$, the graph matching problem is to maximize the similarities between the matched vertices and corresponding edges connected by these vertices. In the following derivation, we use the general notation $\mathcal{G}_1$ and $\mathcal{G}_2$ to obtain a general graph matching formulation.

As defined in \cite{LawlerMS63}, the graph matching problem is a Quadratic Assignment Problem (QAP) . A practical mathematical form is named \emph{Koopmans-Beckmann's} QAP \cite{kbqap}:
\begin{equation}
\label{equ:KBQAP}
\begin{aligned}
& \underset{\MPI}{\text{maximize}}
&& \mathcal{J}(\MPI)=\text{tr}(\MA_1\MPI\MA_2\MPI^\top)+\text{tr}(\MB^\top\MPI),  \\
& \text {s.t.}
&& \MPI\mathbf{1}_{n}= \mathbf{1}_{n}, \MPI^\top\mathbf{1}_{n}= \mathbf{1}_{n},
\end{aligned}
\end{equation}
where $\MPI\in\{0,1\}^{n\times{n}}$ is a permutation matrix that denotes the matching between the vertices of two graphs, $\MA_1\in\MBR^{n\times n}$, $\MA_2\in\MBR^{n\times n}$ are the weighted adjacency matrices of graph $\mathcal{G}_1$ and $\mathcal{G}_2$ respectively, and $\MB\in\MBR^{n\times n}$ is the vertex affinity matrix between $\mathcal{G}_1$ and $\mathcal{G}_2$. $\mathbf{1}_{n}$ denotes an n-dimensional vector with all values to be 1.

\subsection{Reformulation and Convex Relaxation}
\label{sec:qp}
For \emph{Koopmans-Beckmann's} QAP, as $\MPI$ is a permutation matrix, i.e., $\MPI^\top\MPI=\MPI\MPI^\top=\mathbf{I}$. Following~\cite{facgm}, Eq.~\ref{equ:KBQAP} can be rewritten as
\begin{equation}
\begin{aligned}
\mathbf{\Pi}^*
&=\underset{\mathbf{\Pi}}{\arg\min} \ \frac{1}{2}||\mathbf{A_1}\mathbf{\Pi}-\mathbf{\Pi}\mathbf{A_2}||_F^2-\text{tr}(\mathbf{B}^\top\mathbf{\Pi}).
\end{aligned}
\label{K-B}
\end{equation}
%\begin{comment}
%    &=  \underset{\mathbf{\Pi}\in\mathcal{P}}{\arg\max} \ \text{tr}(\mathbf{A_1}\mathbf{\Pi}\mathbf{A_2}\mathbf{\Pi}^\top)+\text{tr}(\mathbf{K}^\top\mathbf{\Pi}) \\
%&=  \underset{\mathbf{\Pi}\in\mathcal{P}}{\arg\max} \ \text{tr}(\mathbf{A_1}\mathbf{\Pi}\mathbf{A_2}\mathbf{\Pi}^\top)  - \frac{1}{2} \text{tr}(\mathbf{A_1}\mathbf{A_1}\mathbf{\Pi}\mathbf{\Pi}^\top) \\
%&\ \ \ \ -\frac{1}{2}\text{tr}(\mathbf{A_2}\mathbf{A_2}\mathbf{\Pi}^\top\mathbf{\Pi})+\text{tr}(\mathbf{K}^\top\mathbf{\Pi}) \\
%&= \underset{\mathbf{\Pi}\in\mathcal{P}}{\arg\max} \ - \frac{1}{2} ||\mathbf{A_1}\mathbf{\Pi}-\mathbf{\Pi}\mathbf{A_2}||_F^2+\text{tr}(\mathbf{K}^\top\mathbf{\Pi}) \\
%\end{comment}
This formulation is more intuitive than that in Eq.~\ref{equ:KBQAP}. For two vertices $i, i' \in \MCG_1$  
and their corresponding vertices $j, j' \in \MCG_2$, the first term in Eq.~\ref{K-B} denotes the difference of the weight of edge $(i, i')$ and $(j, j')$, and the second term denotes the vertex affinities between $i$ and $j$. Then the goal of the optimization is to maximize the vertex affinities between all matched vertices, and minimize the difference of edge weights between all matched edges.
 
It can be proven that the convex hull of the permutation matrix lies in the space of the doubly-stochastic matrix. So, as shown in \cite{aflalo2015convex}, the QAP (Eq.~\ref{K-B}) can be relaxed to its tightest convex relaxation by only constraining the permutation matrix $\mathbf{\Pi}$ to be a double stochastic matrix $\mathbf{X}$, formed as the following QP problem:
\begin{equation}
\mathbf{X}^*=\underset{\mathbf{X}\in\mathcal{D}}{\arg\min} \ \frac{1}{2}||\mathbf{A_1}\mathbf{X}-\mathbf{X}\mathbf{A_2}||_F^2-\text{tr}(\mathbf{B}^\top\mathbf{X}),
\label{QP}
\end{equation}
where $\mathcal{D}=\{\mathbf{X}:\mathbf{X}\mathbf{1}_n= \mathbf{1}_n, \mathbf{X}^\top\mathbf{1}_n=\mathbf{1}_n,\mathbf{X}\geq\mathbf{0}\}$.
\subsection{From Edge Weights to Edge Features}
In the formulation of graph matching above, the element $a_{i,i'}$ in the weighted adjacency matrix $\MA \in\MBR^{n\times n}$ is a scalar denoting the weight on the edge $(i, i')$. To facilitate the application in our MOT problem, we expand the relaxed QP formulation by using an \emph{$l_2$-normalized} edge feature $\mathbf{h}_{i,i'} \in \MBR^d$ instead of the scalar-formed edge weight $a_{i,i'}$ in $\mathbf{A}$. We build a weighted adjacency tensor $\mathbf{H} \in \MBR^{d \times n \times n}$ where $\mathbf{H}^{\cdot, i,i'}$ = $\mathbf{h}_{i, i'}$, i.e., we consider the each dimension of $\mathbf{h}_{i,i'}$ as the element $a_{i,i'}$ in $\mathbf{A}$ and concatenate them along channel dimension. The $\mathbf{H_1}$ and $\mathbf{H_2}$ are the weighted adjacency tensors for $\MCG_1$ and $\MCG_2$, respectively. Then the optimization objective in Eq.~\ref{K-B} can be further expanded to consider the $l_2$ distance between two corresponding \emph{n-d} edge features other than the scalar differences:
\begin{equation}
\begin{aligned}
\mathbf{\Pi}^*
&=\underset{\mathbf{\Pi}}{\arg\min} \ \sum_{c=1}^{d}\frac{1}{2}||\mathbf{H}_1^c\mathbf{\Pi}-\mathbf{\Pi}\mathbf{H}_2^c||_F^2-\text{tr}(\mathbf{B}^\top\mathbf{\Pi}) \\
&=\underset{\mathbf{\Pi}}{\arg\min} \ \sum_{i=1}^{n}\sum_{i'=1}^{n}\sum_{j=1}^{n}\sum_{j'=1}^{n}  \frac{1}{2}||\mathbf{h}_{ii'}\pi_{ij}-\mathbf{h}_{jj'}\pi_{i'j'}||_2^2 \\
&\ \ \ \ -\text{tr}(\mathbf{B}^\top\mathbf{\Pi}) \\
&=\underset{\mathbf{\Pi}}{\arg\min} \ \sum_{i=1}^{n}\sum_{i'=1}^{n}\sum_{j=1}^{n}\sum_{j'=1}^{n}  \frac{1}{2}(\pi_{ij}^2-2\pi_{ij}\pi_{i'j'}\mathbf{h}_{ii'}^\top\mathbf{h}_{jj'} \\
&\ \ \ \ +\pi_{i'j'}^2)-\text{tr}(\mathbf{B}^\top\mathbf{\Pi}),
\end{aligned}
\label{before_edge}
\end{equation}
where $n$ is the number of vertices in graph $\mathcal{G}_1$ and $\mathcal{G}_2$, the subscript $i$ and $i'$ are the vertices in graph $\mathcal{G}_1$ and $j$ and $j'$ are in graph $\mathcal{G}_2$. We reformulate Eq.~\ref{before_edge} as: 
\begin{equation}
\bm{\pi}^*=\underset{\bm{\pi}}{\arg\min} \ 
\bm{\pi}^\top((n-1)^2\mathbf{I}-\mathbf{M})\bm{\pi}-\mathbf{b}^\top\bm{\pi},
\label{edge}
\end{equation}
where $\bm{\pi}=\text{vec}(\MPI)$, $\mathbf{b}=\text{vec}(\MB)$ and $\mathbf{M}\in \MBR^{n^2\times n^2}$ is the symmetric quadratic affinity matrix between all the possible edges in two graphs.

Following the relaxation in Section \ref{sec:qp}, the formulation Eq.~\ref{edge} using edge features can be relaxed to a QP:\\
\begin{equation}
\begin{aligned}
\mathbf{x}^*
&=\underset{\mathbf{x}\in\mathcal{D^{'}}}{\arg\min} \ 
\mathbf{x}^\top((n-1)^2\mathbf{I}-\mathbf{M})\mathbf{x}-\mathbf{b}^\top\mathbf{x}, \\
% &=\underset{\mathbf{x}\in\mathcal{D^{'}}}{\arg\min} \ 
% \frac{1}{2}\mathbf{x}^\top\mathbf{Q}\mathbf{x}+\mathbf{q}^\top\mathbf{x} 
\end{aligned}
\label{finalQP}
\end{equation}
where $ \mathcal{D}^{'}=\{\mathbf{x}:\mathbf{R}\mathbf{x}=\mathbf{1},\mathbf{U}\mathbf{x}\leq \mathbf{1},\mathbf{x}\geq\mathbf{0},\mathbf{R}=\mathbf{1}_{n_2}^\top\otimes{\mathbf{I}_{n_1}},\mathbf{U}=\mathbf{I}_{n_2}^\top\otimes{\mathbf{1}_{n_1}}\}$, $\otimes$ denotes Kronecker product.
\begin{figure}
  \vspace{-0.2cm}
          \centering
           \includegraphics[width=\linewidth]{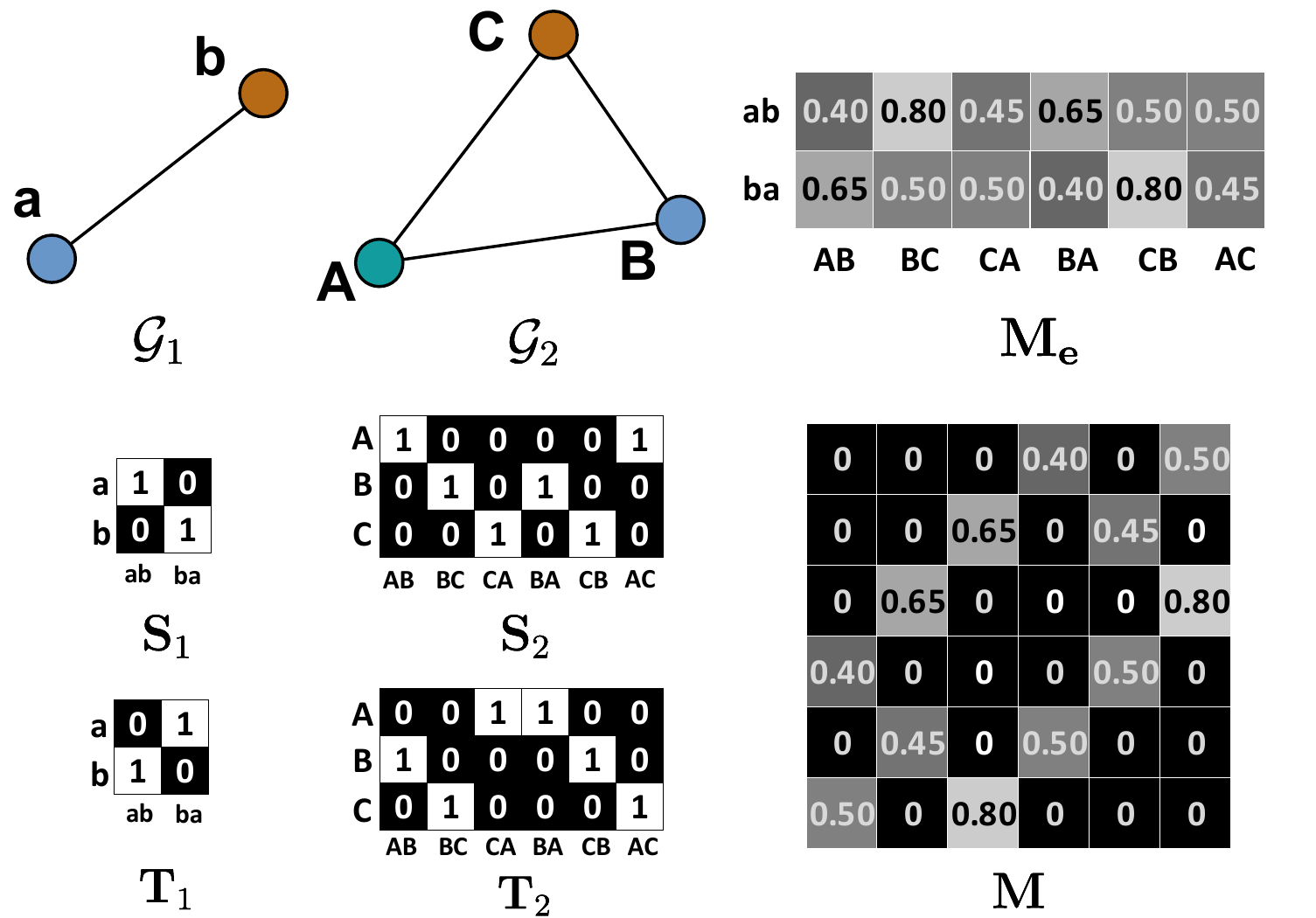}
              
  \vspace{-0.2cm}
             \caption{An example of the derivation from edge affinity matrix $\mathbf{M_e}$ to quadratic affinity matrix $\mathbf{M}$.}
             \label{fig:me2m}
  \end{figure}    

 In the implementation, we first compute the cosine similarity between the edges in $\mathcal{G}_D$ and $\mathcal{G}_T$ to construct the matrix $\mathbf{M_e}\in\MBR^{|\mathcal{E}_D|\times |\mathcal{E}_T|}$. 
 The element of the matrix $\mathbf{M_e}$ is the cosine similarity between edge features $\mathbf{h}_{i,i'}$ and $\mathbf{h}_{j,j'}$ in two graphs:
\begin{align}
  \mathbf{M}_e^{u,v}=\mathbf{h}_{i,i'}^\top\mathbf{h}_{j,j'},
  \label{eq:e2e}
\end{align}
where $e_u=(i,{i'})$ is the edge in $\mathcal{G}_D$ and $e_v=(j,{j'})$ is the edge in $\mathcal{G}_T$. 
 
 And following \cite{zanfir2018deep}, we map each element of matrix $\mathbf{M_e}$ to the \emph{symmetric} quadratic affinity matrix $\mathbf{M}$:
\begin{align}
  \mathbf{M}=(\mathbf{S_D}\otimes\mathbf{S_T})\text{diag}(\text{vec}(\mathbf{M_e}))(\mathbf{T_D}\otimes\mathbf{T_T})^\top,
  \label{eq:me2m}
\end{align}
where $\text{diag}(\cdot)$ means constructing a diagonal matrix by the given vector, $\mathbf{S_D} \in \{0,1\}^{|\MCV_D| \times |\MCE_D|}$ and $\mathbf{S_T} \in \{0,1\}^{|\MCV_T| \times |\MCE_T|}$, whose elements are an indicator function:
\begin{equation}
  \mathbb{I}_s(i,u):= \begin{cases}
    1 & \text{if $i$ is the start vertex of edge $e_u$}, \\
    0 & \text{if $i$ is not the start vertex of edge $e_u$},
  \end{cases}
\end{equation}
$\mathbf{T_D} \in \{0,1\}^{|\MCV_D| \times |\MCE_D|}$ and $\mathbf{T_T} \in \{0,1\}^{|\MCV_T| \times |\MCE_T|}$, whose elements are another indicator function:
\begin{equation}
  \mathbb{I}_t(i',u):= \begin{cases}
    1 & \text{if $i'$ is the end vertex of edge $e_u$}, \\
    0 & \text{if $i'$ is not the end vertex of edge $e_u$}.
  \end{cases}
\end{equation}
An example of the derivation from $\mathbf{M_e}$ to $\mathbf{M}$ is illustrated in Fig.~\ref{fig:me2m}. %In brief, each element in $\mathbf{M_e}$ is mapped to the matrix $\mathbf{M}$ to make the edge affinity matrix a symmetric matrix indexed by the vertex pairs.

Besides, each element in the vertex affinity matrix $\mathbf{B}$ is the cosine similarities between feature $\mathbf{h}_i$ on vertex $i \in \MCV_D$ and feature $\mathbf{h}_j$ on vertex $j\in \MCV_T$:
\begin{align}
  \mathbf{B}_{{i,j}}=\mathbf{h}_{i}^\top\mathbf{h}_{j}
  \label{eq:n2n}
\end{align}
\par
\begin{figure*}[ht]
\vspace{-0.2cm}
        \centering
         \includegraphics[width=\linewidth]{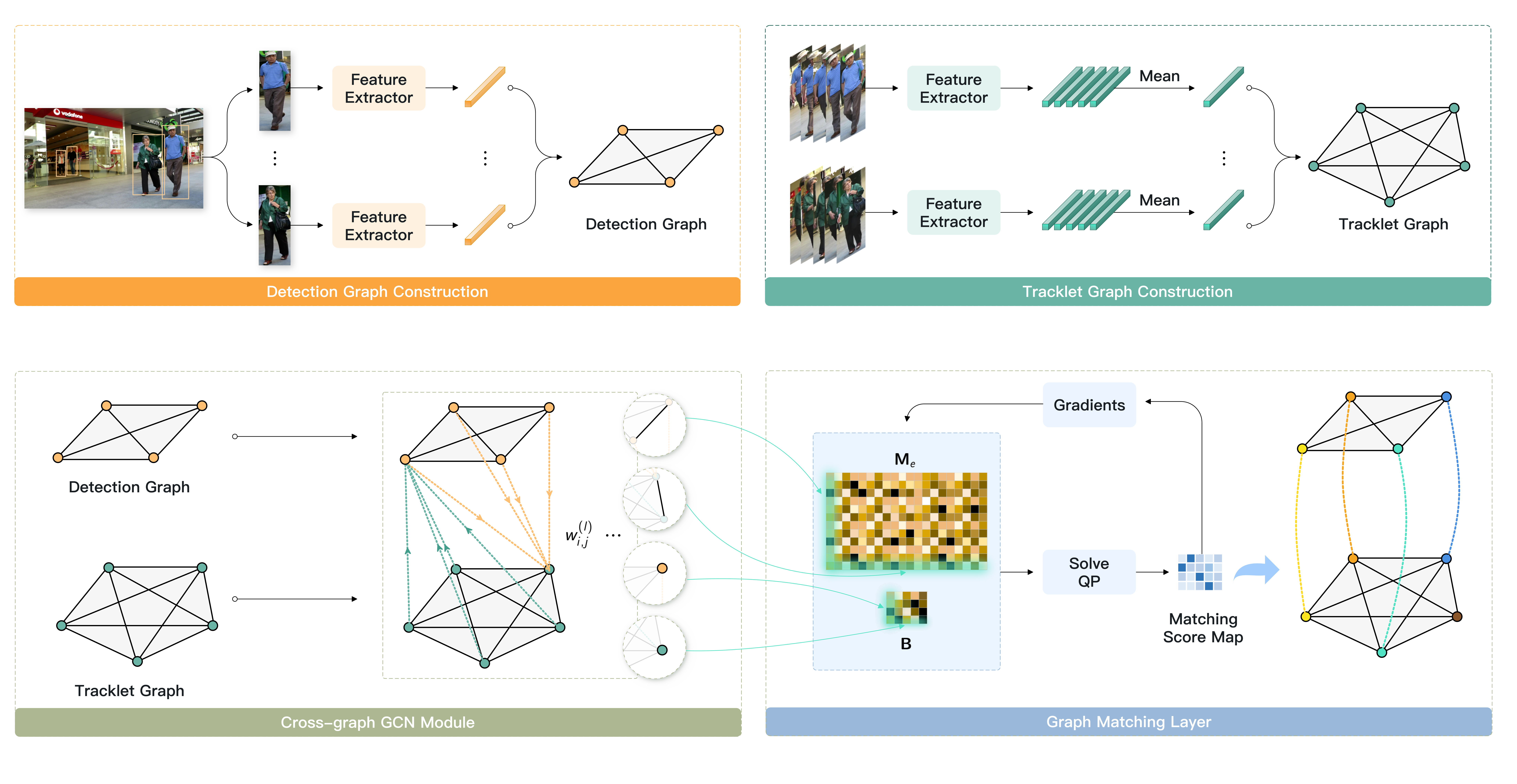}
            
\vspace{-0.5cm}
           \caption{Overview of our method. We first extract features from detections and construct the detection graph using these features. The tracklet graph construction step is similar to the detection graph, but we average the features in a tracklet. Then the cross-graph GCN is adopted to enhance the features. The weight $w_{i,j}$ is from the feature similarity and geometric information. The core of our method is the differentiable graph matching layer built as a QP layer from the formulation in Eq. \ref{finalQP}. The $\mathbf{M}_e$ and $\mathbf{B}$ in the graph matching layer denote the edge affinity matrix from Eq. \ref{eq:e2e} and the vertex affinity matrix from Eq. \ref{eq:n2n} respectively.}
           \vspace{-10pt}
           \label{pipeline}
\end{figure*}    
\section{Graph Matching Network and GMTracker}
In this section, we will describe the details of our Graph Matching Network and our {\gm}. As shown in Fig.~\ref{pipeline}, the pipeline of our Graph Matching Network consists of three parts: (1) feature encoding in detection and tracklet graphs; (2) feature enhancement by cross-graph Graph Convolutional Network (GCN) and (3) differentiable graph matching layer. We will describe these three parts step by step and show how we integrate them into a tracker ({\gm}) in the following.

%\noindent{\bf Graph construction and feature encoding.}
%We construct two graphs, detection graph and tracklet graph. In detection graph, each vertex represents an object in current frame. We encode the appearance feature of the object into vertex. Tracklet graph contains the trajectories information of past frames. We aggregate the features of objects belong to the same tracklet as a vertex in tracklet graph. 
%
%\noindent{\bf Feature propagation using cross-graph GCN.}
%We adopt GCN cross tracklet graph and detection graph to propagation features between detection and tracklet.
%
%%The similarities of appearance and geometric information are encoded on edge between two graphs.
%
%\noindent{\bf Differential graph matching layer.}
%As derived in section \ref{sec:relax}, the graph matching layer can be regarded as a QP layer in our graph matching network.
%% To make graph matching module end-to-end learnable, we redefine a relaxed convex quadratic problem formulation for graph matching problem. 
%We implement the differential graph matching layer following OptNet~\cite{amos2017optnet}.
%\begin{comment}
%{\color{red}
%\noindent{\bf Optimization and loss function.}
%Weighted BCE loss is adopted in our method. Before fed  loss function, the graph matching score matrix is sharpened by a Softmax function with temperature.
%}
%\end{comment}
\subsection{Feature Encoding in Two Graphs}

%\subsection{Feature Encoding}
%\label{sec:fegc}

We utilize a pre-trained ReIDentification (ReID) network followed by a multi-layer perceptron (MLP) to generate the appearance feature $\mathbf{a}_D^{i}$ for each detection $D_i$.
The appearance feature $\mathbf{a}_T^{j}$ of the tracklet $T_{j}$ is obtained by averaging all the appearance features of detections before.

\begin{comment}
{\color{red}We also try other aggregation function like moving average, and find simple average can yield better performance.} 
\end{comment}
%Note that we do not associate edges in $\MCG_D^t$ and $\MCG_T^t$ with edge features currently. We will do this after cross-graph GCN.
\begin{comment}
\begin{equation}
  \begin{aligned}
    \mathbf{a}_T^{i,t}=\frac{1}{|K|}\sum_{k\in K}{\rm MLP_{e}}\comp {\rm Enc}(\mathbf{I}_{p,i}^k)
  \end{aligned}
\end{equation}
where $K=\{k|k<t\}$. 
\end{comment}
%We also try other feature aggregation methods, such as moving average in the whole tracklet, moving average in the last few frames, etc. 
%Comparing these methods, we finally choose the simplest but effective way, averaging the appearance feature during the whole tracklet. 
%For the comparison between these methods, see Table \ref{tbl:traagg}.
\par
\begin{comment}
Then, two complete graphs are constructed. Let $\mathcal{G}_T^t=(\mathcal{V}_T,\mathcal{E}_T)$ be a tracklet graph. The vertex $n_i$ in vertex set $\mathcal{V}_T$ corresponds to the tracklet $T_i^t$. The edges in edge set $\mathcal{E}_T=\{e_{i,j}\}$ are the fully connections between two arbitrary vertices in graph $\mathcal{G}_T^t$. Each vertex feature $\mathbf{h}_i$ of $n_i$ is the appearance feature embedding of the tracklet and the edge feature $\mathbf{h}_{i,j}$ on the edge $e_{i,j}$ contains two vertex features, as
\begin{equation}
  \begin{aligned}
    \mathbf{h}_i^{(0)}=\mathbf{a}_T^{i,t}, \ \ \mathbf{h}_{i,j}^{(0)}=\bm{[}\mathbf{a}_T^{i,t},\mathbf{a}_T^{j,t}\bm{]}
\end{aligned}
\end{equation}
where, $\bm{[\cdot]}$ denotes concatenation operation.
\par
The detection graph $\mathcal{G}_D^t=(\mathcal{V}_D,\mathcal{E}_D)$ is similar to the tracklet graph. Each vertex includes a detection appearance feature embedding, and each edge is the combination of two connected vertex features, as
\begin{equation}
  \begin{aligned}
    \mathbf{h}_p^{(0)}=\mathbf{a}_D^{p,t},\ \  \mathbf{h}_{p,q}^{(0)}=\bm{[}\mathbf{a}_D^{p,t},\mathbf{a}_D^{q,t}\bm{]}.
\end{aligned}
\end{equation}
\end{comment}

\subsection{Cross-Graph GCN}
\label{sec:gcn}
 Similar to \cite{braso2020learning, ma2019deep, Weng2020_GNN3DMOT}, we only adopt a GCN module between the graph $\MCG_D$ and graph $\MCG_T$ to enhance the feature, and thus it is called Cross-Graph GCN. 

 The initial vertex features on detection graph and tracklet graph are the appearance features on the vertices, i.e., let $\Mh_i^{(0)}=\mathbf{a}_D^{i}$ and $\Mh_j^{(0)}=\mathbf{a}_T^{j}$.
 \begin{comment}
 In the following, we describe how we enhance the feature of a detection vertex by aggregating the features from the tracklet graph. Similar procedure is also applied to the tracklet vertices.
 \end{comment}
Let $\Mh_i^{(l)}$ and $\Mh_j^{(l)}$ be the feature of vertex $i \in \MCG_D$ and vertex $j \in \MCG_T$ in the $l$-th propagation, respectively. 
\begin{comment}
Then the feature update for vertex $i$ can be formulated as:
\begin{align}
	&\Mm_{i}^{(l)} = \MCA(\{w_{i,j}^{(l)}\Mh_j^{(l)} \mid j \in \MCG_T\}), \\
	&\Mh_{i}^{(l + 1)} = \MCF(\Mh_i^{(l)}, \Mm_{i}^{(l)}),
\end{align}
where $w_{i,j}$ is a weight coefficient, $\MCA(\cdot)$ and $\MCF(\cdot)$ are the aggregation and the transformation function, respectively.
\end{comment}
We define the aggregation weight coefficient $w_{i, j}^{(l)}$ in GCN as the appearance and geometric similarity between vertex $i$ and vertex $j$:
\begin{equation}
w_{i,j}^{(l)} = \cos(\Mh_i^{(l)}, \Mh_j^{(l)}) + {\rm{IoU}}(\mathbf{g}_i, \mathbf{g}_j)
\end{equation}
where $\cos(\cdot, \cdot)$ means the cosine similarity between input features and $\rm{IoU(\cdot, \cdot)}$ denotes the Intersection over Union of two bounding boxes. For a detection vertex $i$, $\mathbf{g}_i$ is the corresponding detection bounding box defined in Section \ref{sec:construct}. As for a tracklet vertex $j$, we estimate the bounding box $\mathbf{g}_j$ in current frame $t$ by Kalman Filter \cite{kalman1960new} motion model with a constant velocity. Note that we only consider the appearance feature similarity in weight $w_{i,j}$ when the camera moves, since the motion model cannot predict reliable future positions in these complicated scenes.

We use summation as the aggregation function, i.e., $\Mm_{i}^{(l)} = \sum_{j \in \MCG_T} w_{i,j}^{(l)}\Mh_j^{(l)}$ and the vertex features are updated by:
\begin{equation}
\Mh_i^{(l + 1)} = {\rm{MLP}}(\Mh_i^{(l)} + \frac{\norm{\Mh_i^{(l)}}_2\Mm_{i}^{(l)}}{\norm{\Mm_{i}^{(l)}}_2}),
\end{equation}
where we adopt message normalization proposed in \cite{li2020deepergcn} to stabilize the training.

We apply $l_2$ normalization to the final features after cross-graph GCN and denote it as $\Mh_i$. Then we use $\Mh_i$ as the feature of vertex $i$ in graph $\MCG_D$, and construct the edge feature for edge $(i, i')$ with $\Mh_{i,i'} = l_2([\Mh_i, \Mh_{i'}])$, where $[\cdot]$ denotes concatenation operation. The similar operation is also applied to the tracklet graph $\MCG_T$. In our implementation, we only apply GCN once.

\subsection{Differentiable Graph Matching Layer}
\begin{comment}
\footnote{A key issue of the section is how does M connected to Eqn6? What is the meaning of Graph matching in the context of MOT? These points need to be clarified before going into the details.}
\end{comment}
After enhancing the vertex features and constructing the edge features on graph $\MCG_D$ and $\MCG_T$, we meet the core component of our method: the differentiable graph matching layer. By optimizing the QP in Eq.~\ref{finalQP} from quadratic affinity matrix $\mathbf{M}$ and vertex affinity matrix $\mathbf{B}$, we can derive the optimal matching score vector $\mathbf{x}$ and reshape it back to the shape $n_d \times n_t$ to get the matching score map $\mathbf{X}$.

Since we finally formulate the graph matching problem as a QP, we can construct the graph matching module as a differentiable QP layer in our neural network. Since KKT conditions are the necessary and sufficient conditions for the optimal solution $\mathbf{x}^*$ and its dual variables, we could derive the gradient in backward pass of our graph matching layer based on the KKT conditions and implicit function theorem, which is inspired by OptNet~\cite{amos2017optnet}. Please refer to the appendix for the detailed derivation of the gradients in graph matching layer.
% \footnote{Put the details in supplemental material} 
In our implementation, we adopt the qpth library~\cite{amos2017optnet} to build the graph matching module.
% Besides solving the QP, in the neural network, the differential QP layer needs back propagation. According to the matrix
% differential calculus, the derivative of the solution can be obtained from KKT conditions, which is the sufficient and necessary conditions for the optima. In our graph matching module, the gradients required in back propagation, can be formulated ~\cite{amos2017optnet} as 
% \begin{equation}
%   \begin{aligned}
%     \nabla_\mathbf{Q} \ell = \frac{1}{2}(d_x x^\top + xd_x^\top),
%     \nabla_\mathbf{q} \ell = d_x 
%   \end{aligned}
%   \label{eq:grads}
% \end{equation}
% where, $\ell$ is the loss function.
In the inference stage, to reduce the computational cost and accelerate the algorithm, we solve the QP using the CVXPY library~\cite{diamond2016cvxpy} only for forward operation.
\begin{comment}
Following FGM~\cite{}, the edge affinity matrix between two graphs, can be formed as 
\begin{equation}
  \mathbf{M}=(\mathbf{G_2}\otimes\mathbf{G_1})[\text{vec}(\mathbf{M_e})](\mathbf{H_2}\otimes\mathbf{H_1})^T
\end{equation}

We can adopt the methods in OptNet~\cite{amos2017optnet} and Cvxpylayer~\cite{AgrawalABBDK19}, regarding the graph matching as a QP layer, which can be end-to-end trained. In the implementation, we use qpth library.
\par
\noindent{\bf Graph matching using vertex and edge features.} \ 
In this part, we expand the relaxed QP Eq.\ref{QP} formulation using vertex and edge features.

\noindent{\bf Solving the QP and back propagation.} \ 
In this part, we describe how to solve the QP and how to compute gradients in the process of back propagation.
\begin{equation}
  \begin{aligned}
    \nabla_Q \ell &= \frac{1}{2}(d_z z^T + zd_z^T) &
    \nabla_q \ell &= d_z \\
  \end{aligned}
  \label{eq:grads}
\end{equation}
\end{comment}
\par
For training, we use weighted binary cross entropy Loss: 
\begin{equation}
  \begin{aligned}
    \mathcal{L}=\frac{-1}{n_dn_t}\sum_{i=1}^{n_d}\sum_{j=1}^{n_t}\ ky_{i,j}\log(\hat{y}_{i,j})+(1-y_{i,j})\log(1-\hat{y}_{i,j}),
\end{aligned}
\end{equation}
where $\hat y_{i,j}$ denotes the matching score between detection $D_i$ and tracklet $T_j$, and $y_{i,j}$ is the ground truth indicating whether the object belongs to the tracklet. $k=(n_t-1)$ is the weight to balance the loss between positive and negative samples. Besides, due to our QP formulation of graph matching, the distribution of matching score map $\mathbf{X}$ is relatively smooth. We adopt softmax function with temperature $\tau$ to sharpen the distribution of scores before calculating the loss:
\vspace{-2pt}
\begin{equation}
\hat{y}_{i,j} = {\rm Softmax}(x_{i,j},\tau)=\frac{e^{x_{i,j}/\tau}}{\sum_{j=1}^{n_t}e^{x_{i,j}/\tau}},
\label{eq:softmax}
\vspace{-2pt}
\end{equation}
where $x_{i,j}$ is the original matching score in score map $\mathbf{X}$.
\subsection{Inference Details}
\label{sec:tarcker}
%After training, the proposed pipeline can be straightly adopted as an online tracker. Graph construction, feature encoding and graph matching procedures are exactly the same as the training stage.
Due to the continuous relaxation, the output of the QP layer may not be binary. To get a valid assignment, we use the greedy rounding strategy to generate the final permutation matrix from the predicted matching score map, i.e., we match the detection with the tracklet with the maximum score. 
After matching, like DeepSORT \cite{wojke2017simple}, we need to handle the born and death of tracklets. We filter out the detection if it meets one of the three criteria: 1) All the appearance similarities between a detection and existing tracklets are below a threshold $\sigma$. 2) It is far away from all tracklets. We set a threshold $\kappa$ as the Mahalanobis distance between the predicted distribution of the tracklet bounding box by the motion model and the detection bounding box in pixels, called motion gate. 3) The detection bounding box has no overlap with any tracklets. Here, besides the Kalman Filter adopted to estimate the geometric information in Section \ref{sec:gcn}, we apply an Enhanced Correlation Coefficient (ECC) \cite{ecc} in our motion model additionally to compensate the camera motion. Besides, we apply the IoU association between the filtered detections and the unmatched tracklets by Hungarian algorithm to compensate some incorrect filtering.
% \footnote{Confused, since one of the criteria is no overlap with any tracklet. How to do this association? Hungarian again?}
 Then the remaining detections are considered as a new tracklet. We delete a tracklet if it has not been updated since $\delta$ frames ago, called \emph{max age}.  
\begin{comment}
After matching, We follow DeepSORT \cite{wojke2017simple} to handle the born and death of tracklets. We start a new tracklet if a detection meets one of the three criteria: 1) all the appearance similarities between a detection and existing tracklets are below a threshold $\sigma$. 2) it is far away from all tracklets, i.e. its center location is out of motion gate $\kappa$.\footnote{This one is not clear to me.} 3) the bounding box has no overlap with any tracklets. The motion model we used \footnote{What for?}is Kalman Filter, but for the video taken by the moving camera, we apply an Enhanced Correlation Coefficient (ECC) \cite{ecc} model additionally to compensate the camera motion.\footnote{Then why not apply it to the construction of edge weight?} For the unmatched detections\footnote{Is it right?}, we apply IoU matching between the filtered detections\footnote{What does filtered mean?} and the unmatched tracklets.\footnote{Then when do we delete a tracklet?}
\end{comment}
\vspace{-3pt}
\section{Experiments}

\subsection{Datasets}
%We carry out all experiments on MOTChallenge benchmark, which is a challenging multiple pedestrian tracking dateset. MOTChallenge benchmark includes four main datasets, 2DMOT15~\cite{leal2015motchallenge}, MOT16~\cite{milan2016mot16}, MOT17~\cite{milan2016mot16} and MOT20~\cite{dendorfer2020mot20}. 
We carry out all experiments on MOT16~\cite{milan2016mot16} and MOT17~\cite{milan2016mot16} benchmark.
The videos in this benchmark were taken under various scenes, light conditions and frame rates. Occlusion, motion blur, camera motion and distant pedestrians are also crucial problems in this benchmark. Among all the evaluation metrics, Multiple Object Tracking Accuracy (MOTA)~\cite{kasturi2008framework} and ID F1 Score (IDF1)~\cite{ristanieccvw2016} are the most general metrics in the MOT task. Since MOTA is mostly dominated by the detection metrics false positive and false negative, and our graphing matching method mainly tries to tackle the associations between detected objects, we pay more attention to IDF1 than the MOTA metric. %We report our results on MOT16 and MOT17.
\begin{table*}
    \begin{center}
      \begin{tabular}{ccccc|cc|ccccc}
        \toprule
        GM &App. Enc. &GCN &Geo &Inter. & IDF1 $\uparrow$& MOTA $\uparrow$& MT $\uparrow$& ML $\downarrow$& FP $\downarrow$& FN $\downarrow$& ID Sw. $\downarrow$\\
        \midrule
        &&& & &  68.1 &  62.1  &  556 &  371  &  1923  &  124480  &  1135  \\
        \checkmark&&& &   &  70.0  &  62.3 & 555 &  374  &  1735  & 124292 &1128\\
        \checkmark&&&\checkmark &   &70.2 &  62.2  &  555 & 374 & 1744  & 124301  &  1140 \\
        \checkmark&\checkmark&& & &70.4 & 62.3  &  554 &  375  &  1741  & 124298  & 1058  \\
        \checkmark&\checkmark&\checkmark& &  &70.6 &  62.2  &  556 &  374  &  1748  &  124305  &  1399  \\
        \checkmark&\checkmark&\checkmark&\checkmark & &71.5 & 62.3  &  555 &  375  & 1741  &  124298  & 1017  \\
        \midrule
        &&&  &\checkmark &  68.9 &  62.9  & 678 &  361  &  11440  &  112853  &  723  \\
        \checkmark&&&   &\checkmark  &71.6 &  64.0  &  669 &  365  &  7095  &  113392  & 659  \\
        \checkmark&&&\checkmark& \checkmark   &71.7 & 64.0  & 666 &  364  & 6816  & 113778  & 724 \\
        \checkmark&\checkmark&&  &\checkmark  &72.0 &  64.2  & 671 &  368  &  7701  &  112370  &  627  \\
        \checkmark&\checkmark&\checkmark&    &\checkmark  &72.1 &  63.3  &  676 &  364  &  10888  &  111869  & 716  \\
        \checkmark&\checkmark&\checkmark&\checkmark &\checkmark  &73.0 &  63.8  & 672 &  361  &  9579  & 111683  &  570  \\
        \bottomrule
      \end{tabular}
      \end{center}
      \vspace{-0.2cm}
    \caption{Ablation studies on different proposed components on MOT17 \emph{val} set.}
    \vspace{-10pt}
  \label{tbl-ablation}
  \end{table*}
\vspace{-10pt}
\subsection{Implementation Details}
\noindent\textbf{Training.}
Following other MOT methods \cite{braso2020learning,hornakova2020lifted}, we adopt Tracktor~\cite{bergmann2019tracking} to refine the public detections. For the ReID network used for feature extraction, we use a ResNet50 \cite{he2016deep} backbone followed by a global average pooling layer and a fully connected layer with 512 channels. We further normalize the output feature with the $l_2$ normalization. We pre-train the ReID network on Market1501~\cite{market_dataset}, DukeMTMC~\cite{ristanieccvw2016} and CUHK03~\cite{cuhk03_dataset} datasets jointly, following the setting of \cite{braso2020learning}. The parameters of the ReID network will be frozen after pre-training. Then we add two trainable fully connected layers with 512 channels to get appearance features.
Our implementation is based on PyTorch~\cite{paszke2019pytorch} framework. We train our model on an NVIDIA RTX 2080Ti GPU. Adam~\cite{kingma2014adam} optimizer is applied with $\beta_1=0.9$ and $\beta_2=0.999$. The learning rate is  5$\times$10$^{-5}$ and weight decay is 10$^{-5}$. The temperature $\tau$ in Eq. \ref{eq:softmax} is 10$^{-3}$. 

\noindent\textbf{Inference.}
Our inference pipeline mostly follows DeepSORT \cite{wojke2017simple}, except that we use general graphing matching instead of bipartite matching for association. As in DeepSORT, we set the motion gate $\kappa$ as 9.4877, which is at the 0.95 confidence of the inverse $\chi^2$ distribution. The feature similarity threshold $\sigma$ is set to 0.6 in the videos taken by the moving camera, and 0.7 when we use geometric information in the cross-graph GCN module for videos taken by the static camera. The \emph{max age} $\delta$ is 100 frames.
\par
\noindent\textbf{Post-Processing.} To compare with other state-of-the-art offline methods, we perform a linear interpolation within the tracklet as post-processing to compensate the missing detections, following \cite{braso2020learning,hornakova2020lifted}. This effectively reduces the false negatives introduced by upstream object detection algorithm.
% In ablation study, we compare our proposed components under the experiment settings both with and without the interpolation. In Table \ref{tab:mot}, we show both to compare with other state-of-the-art methods separately.
% \input{tables/big-ablation.tex}

\subsection{Ablation Study}
\label{sec:ablation}

\begin{table}
    \begin{center}
      \begin{tabular}{cc|cc}
        \toprule
        Train w/ GM & Inference w/ GM & IDF1 & MOTA \\
        \midrule
        & &69.5 &62.1\\
        &\checkmark &70.2 &62.3\\
        \checkmark &\checkmark &71.5 &62.3\\
        \bottomrule
      \end{tabular}
      \end{center}
      \vspace{-0.2cm}
    \caption{Ablation study on the graph matching layer.}
    \vspace{-13pt}
  \label{tab-rmgmlayer}
  \end{table}

We conduct ablation studies of the proposed components in our method on the MOT17 dataset. Following \cite{braso2020learning}, we divide the training set into three parts for three-fold cross-validation, called MOT17 \emph{val} set, and we conduct the experiments under this setting both in the ablation study section and the discussions section. We ablate each component we propose: (i) graph matching module built as a QP layer (GM); (ii) MLP trained on MOT dataset to refine the appearance features (App. Enc.); (iii) the cross-graph GCN module (GCN) with and without using geometric information (Geo); (iv) the linear interpolation method between the same object by the time (Inter.).

As shown in Table \ref{tbl-ablation}, compared with the DeepSORT baseline (the first row), which associates the detections and the tracklets by Hungarian Algorithm, the graph matching method gets a gain of 1.9 IDF1 without interpolation, and a gain of 2.7 IDF1 and 1.1 MOTA with the linear interpolation. The results show the effectiveness of the second-order edge-to-edge information. 
% Because DeepSORT filters the unconfirmed tracklets and keeps the unmatched tracklets for one more frame, it reduces ID switch at the cost of increasing FP and FN. After adopting linear interpolation, the number of ID Switch of our method reduces substantially. However, ID Switch of DeepSORT barely keeps unchanged.

Appearance feature refinement and GCN improve about 0.6 IDF1 compared to the untrained model. Geometric information provides about 1.0 additional gain on IDF1, which highlights the importance of geometric information in the MOT task. Finally, compared with the baseline, our method achieves about 3.4 and 0.2 improvements on IDF1 metric and MOTA metric, respectively. With interpolation, the gain becomes even larger: about 4.1 improvements on IDF1 and 0.9 on MOTA.

As shown in Table \ref{tab-rmgmlayer}, we get the gain of 1.3 and 2.0 IDF1 compared with only removing the graph matching layer in training stage and in both training and inference stage, respectively. The results demonstrate the effectiveness of our differentiable graph matching layer and the importance of training all components in our tracker jointly.

\subsection{Discussions}
In this part, we discuss two main design choices of our method on MOT17 \emph{val} set. When we construct the tracklet graph, there are some different intra-tracklet feature aggregation methods. Moreover, how to create and delete a tracklet is important for an online tracker.

\noindent{\bf Intra-tracklet feature aggregation.} In the tracklet graph $\MCG_T$, each vertex represents a tracklet. And the vertex feature $\mathbf{a}_T^{j}$ is the aggregation of the appearance features of all detections in tracklet $T_j$. Here, we compare several aggregation methods, including mean, moving average and only using the last frame of the tracklet. The results are shown in Table \ref{tbl:traagg}. The IDF1 is 0.9 lower when only using the last frame of the tracklet. The results also reveal that when we utilize all the frame information, no matter using the simple average or the moving average, their impact is not significant. To make our method simple and effective, we finally use the simple average method to aggregate the appearance features within a tracklet.
\begin{table}
    \center
      \begin{tabular}{l|cc}
        \toprule
        \multicolumn{1}{c|}{Methods} & IDF1 & MOTA  \\
        \midrule
        Last Frame  &  69.1 &  62.3  \\
        Moving Average $\alpha=0.5$ &  69.9&  62.3  \\
        Moving Average $\alpha=0.8$ &  70.1&  62.3 \\
        Mean  &  70.0 &  62.3    \\
        \bottomrule
      \end{tabular}
    \caption{Ablation studies on different intra-tracklet feature aggregation methods.}
    \vspace{-5pt}
  \label{tbl:traagg}
  \end{table}
\begin{comment}
\begin{figure}
    \includegraphics[width=\columnwidth]{figures/thr.pdf}
    \caption{Results on IDF1, FP, FN and ID Switch metrics under different threshold $\sigma$ of the feature similarity to create a new tracklet.}
    \vspace{-0.2cm}
    \label{fig:thr}
    \vspace{-3pt}
\end{figure}    
\end{comment}

\noindent{\bf Tracklet death strategies.} 
\begin{comment}
In most of the online tracking methods, one of the core strategies is how to create and delete a tracklet. In our GMTracker, we mostly follow DeepSORT, but we also make some improvements to make these strategies more suitable for our approach, as described in Section \ref{sec:tarcker}.
 Among the three criteria to create a new tracklet, we find that the threshold $\sigma$ is the most sensitive hyperparameter in our method. We conduct experiments with different $\sigma$, and its influence on IDF1, FP, FN and ID Switch is shown in Fig. \textcolor{red}{A} in the appendix.
%  \ref{fig:thr}. 
\end{comment}
\begin{comment}
\begin{table}
    \center
      \begin{tabular}{c|cc}
        \hline

        Max age (frames) & IDF1 & MOTA  \\
        \hline
        30 &  N/A &  N/A   \\
        50  &  69.3 & 62.3   \\
        80 &  N/A &  N/A    \\
        100  &  70.0 & 62.3   \\
        \hline
      \end{tabular}
    \caption{Comparison between diffete}
  \label{tbl-ablation2}
  \end{table}
\end{comment}
  \begin{table}
    \center
    \resizebox{\columnwidth}{!}{
      \begin{tabular}{c|ccccc}
        \toprule
        Max age (frames) &30 &50& 80 &100 &150\\
        \midrule
        Hungarian Algorithm  & 67.2 &67.9  &68.1 &68.1 &67.3\\
        Graph Matching  &  68.0 & 69.3 &69.7 &70.0 &70.0\\
        \bottomrule
      \end{tabular}}
    \caption{The influence of \emph{max age} $\delta$ on IDF1 .}
    \vspace{-14pt}
  \label{tbl-ablation2}
  \end{table}
% With a higher $\sigma$, the number of false positive (FP) will decrease and the false negative (FN) will increase. However, the IDF1 and ID Switch have 
As for removing a trajectory from association candidates, our basic strategy is that if the tracklet has not been associated with any detections in $\delta$ frames, the tracklet will be removed and not be matched any more.
Table \ref{tbl-ablation2} shows that larger \emph{max age} $\delta$, which means more tracklet candidates, yields better IDF1 score. It shows the effectiveness of our method from another aspect that our GMTracker can successfully match the tracklets disappeared about five seconds ago. On the contrary, when the \emph{max age} increases to 150 frames, the IDF1 will drop 0.8 using Hungarian algorithm, which indicates our graph matching can deal with long-term tracklet associations better. %The results show that there is an obvious gap between Hungarian algorithm and our graph matching when the \emph{max age} increases. 
\subsection{Comparison with State-of-the-Art Methods}
We compare our GMTracker with other state-of-the-art methods on MOT16 and MOT17 test sets. As shown in Table \ref{tab:mot}, when we apply Tracktor~\cite{bergmann2019tracking} to refine the public detection, the \emph{online} \gm\  achieves 63.8 IDF1 on MOT17 and 63.9 IDF1 on MOT16, outperforming the other online trackers. To compare with CenterTrack~\cite{zhou2020tracking}, we use the same detctions, called GMT\_CT, and the IDF1 is 66.9 on MOT17 and 68.6 on MOT16.
% compared with other \emph{online} methods, we improve the state-of-the-art on the IDF1 metric by 6.0 and 5.7 on MOT17 and MOT16 datasets, respectively. 
With the simple linear interpolation, called GMT\_simInt in Table \ref{tab:mot}, we also outperform the other \emph{offline} state-of-the-art trackers on IDF1. With exactly the same visual inter- and extrapolation as LifT~\cite{hornakova2020lifted}, called GMT\_VIVE in Table~\ref{tab:mot}, the MOTA is comparable with LifT. After utilizing the CenterTrack detections and linear interpolation, the GMTCT\_simInt improves the SOTA on both MOT16 and MOT17 datasets.
%  The online tracker  refines the public detections based on a better detector CenterNet~\cite{zhou2019objects}. To compare with it, we use the same detections as CenterTrack, called GMT\_CT in Table~\ref{tab:mot}, and we improve the IDF1 by 7.3 on the MOT17 test set. CenterTrack does not report the results on the MOT16 test set. But compared with other methods, the results on MOT16 test set also show the effectiveness of our method. 
In appendix, we report more detailed performance on other metrics, e.g., HOTA~\cite{luiten21hota}.
\begin{table}
    \centering

        \tabcolsep=0.06cm
        \scalebox{0.74}{
        \begin{tabular}{l |c|c c c c c }
         \toprule
         Methods &Refined Det& IDF1 $\uparrow$ & MOTA $\uparrow$  & FP $\downarrow$ & FN $\downarrow$ & IDS $\downarrow$ \\ [0.5ex] 
         \midrule
         \multicolumn{7}{c}{MOT17} \\
         \midrule
         GNMOT (O$^*$) \cite{Li_2020_WACV} &- &47.0&50.2   &29316 &246200 &5273 \\
         FAMNet (O) \cite{chu2019famnet}&- &48.7&52.0  &14138 &253616 &3072 \\
         JBNOT (O$^*$) \cite{henschel2019multiple}&- &50.8&52.6  &31572 &232659 &3050 \\     
         Tracktor++ (O)  \cite{bergmann2019tracking}&Tracktor & 52.3& 53.5 & 12201 & 248047 & 2072 \\
         Tracktor++v2 (O)  \cite{bergmann2019tracking}&Tracktor& 55.1 & 56.3  & 8866 & 235449 & 1987 \\
         GNNMatch (O) \cite{papakis2020gcnnmatch}&Tracktor  &56.1&57.0 &12283 	&228242 &1957 \\
         GSM\_Tracktor (O) \cite{liugsm}&Tracktor &57.8&56.4&14379 &230174 &\textbf{1485} \\
         CTTrackPub (O) \cite{zhou2020tracking} &CenterTrack &59.6&61.5   &14076 &200672 &2583 \\
         \bf{GMTracker(Ours)} (O) &Tracktor& 63.8 & 56.2   &\textbf{8719} & 236541 & 1778  \\
         \bf{GMT\_CT(Ours)} (O) &CenterTrack& \textbf{66.9}& \textbf{61.5}  & 14059 & \textbf{200655} & 2415  \\
         \midrule
         TPM \cite{peng2020tpm} &- &52.6 &54.2   &13739 &242730 &1824 \\
         eTC17 \cite{wang2019exploit}&-  &58.1 &51.9&36164 &232783 &2288 \\
         MPNTrack~\cite{braso2020learning} &Tracktor& 61.7 & 58.8 & 17413 &213594 &1185 \\
         Lif\_TsimInt~\cite{hornakova2020lifted}&Tracktor&65.2&58.2  &16850 &217944 &\textbf{1022} \\
         LifT~\cite{hornakova2020lifted}&Tracktor &65.6 &60.5 &14966 &206619 &1189\\
         \bf{GMT\_simInt (Ours)} &Tracktor&  65.9& 59.0 &  20395 & 209553 & 1105  \\
         \bf{GMT\_VIVE (Ours)}&Tracktor &65.9 & 60.2 &\textbf{13142} &209812 &1675\\
         \bf{GMTCT\_simInt (Ours)}&CenterTrack &\textbf{68.7} &\textbf{65.0}&18213 &\textbf{177058} &2200\\
         \midrule
         \multicolumn{7}{c}{MOT16} \\
         \midrule
         Tracktor++v2 (O)  \cite{bergmann2019tracking} &Tracktor&54.9 &56.2 &2394 	&76844 	&617 \\
         GNNMatch (O) \cite{papakis2020gcnnmatch} &Tracktor&55.9  &56.9&3235 	&74784 	&564 \\
         GSM\_Tracktor (O)\cite{liugsm} &Tracktor &58.2 &57.0   &4332 &73573 &\textbf{475} \\
         \bf{GMTracker(Ours)} (O)  &Tracktor&  63.9& 55.9 & \textbf{2371}& 77545 & 531  \\
         \bf{GMT\_CT (Ours)} (O)&CenterTrack &  \textbf{68.6} &\textbf{ 62.6}  &  5104 &  \textbf{62377} & 787  \\
         \midrule
         TPM \cite{peng2020tpm} &- &47.9  &51.3	&\textbf{2701} &85504 &569 \\ 
         eTC \cite{wang2019exploit} &-  &56.1&49.2 	&8400 	&83702 &606 \\
         MPNTrack~\cite{braso2020learning}&Tracktor  & 61.7& 58.6   & 4949 &70252 &354 \\
         Lif\_TsimInt~\cite{hornakova2020lifted} &Tracktor &64.1&57.5 &4249 &72868 &\textbf{335} \\
         LifT~\cite{hornakova2020lifted}&Tracktor&64.7 &61.3  &4844 &65401 &389\\
         \bf{GMT\_simInt (Ours)} &Tracktor&  66.2 & 59.1   &  6021 &  68226 & 341  \\
         \bf{GMT\_VIVE (Ours)}&Tracktor&66.6 &61.1 &3891 &66550 &503 \\
         \bf{GMTCT\_simInt (Ours)}&CenterTrack  &\textbf{70.6} &\textbf{66.2} &6355 &\textbf{54560} &701\\
         \bottomrule
        \end{tabular}}
    
    \caption{Comparison with state-of-the-art methods on MOT16 and MOT17 \emph{test} set. (O) denotes online methods. (O$^*$) denotes near-online methods.}
    \vspace{-15pt}
    \label{tab:mot}
    \end{table}
\begin{comment}
Furthermore, our differentiable graph matching module can be easily adopted in other classical or deep learning methods to replace the traditional bipartite matching and improve their performance. 
\vspace{-5pt}
\end{comment}
\vspace{-5pt}
\section{Conclusion}

In this paper, we propose a novel learnable graph matching method for multiple object tracking task, called GMTracker. Our graph matching method focuses on the relationship between tracklets and detections. Taking the second-order edge-to-edge similarity into account, our tracker is more accurate and robust in the MOT task, especially in crowded videos. To make the graph matching module end-to-end differentiable, we relax the QAP formulation into a convex QP and build a differentiable graph matching layer in our Graph Matching Network. The experiments of ablation study and comparison with other state-of-the-art methods both show the effectiveness of our method.
\vspace{-5pt}
\section*{Acknowledgements}
\vspace{-5pt}
\noindent{This work was supported in part by the National Key R\&D Program of China(No. 2018YFB1004602), the National Natural Science Foundation of China (No. 61836014, No. 61773375). The authors would like to thank Roberto Henschel for running their post-processing code for us.}

\def\comp{\ensuremath\mathop{\scalebox{.6}{$\circ$}}}
\section*{\LARGE Appendix}
\vspace{5pt}
% \def\paperTitle{Learnable Graph Matching: Incorporating Graph Partitioning with Deep Feature Learning for Multiple Object Tracking \\ {\normalfont Supplementary Material}}
% % %%%%%%%%% TITLE
% % \title{Supplementary Material} \\
% \title{\paperTitle}
% % \LARGE{Learnable Graph Matching: Incorporating Graph Partitioning with Deep Feature Learning for Multiple Object Tracking}
% \author{
%         Jiawei He$^{1,3}$ \quad 
%         Zehao Huang$^{2}$ \quad 
%         Naiyan Wang$^{2}$ \quad 
%         Zhaoxiang Zhang$^{1,3,4}$ \\
%         $^{1}$ Institute of Automation, Chinese Academy of Sciences (CASIA)\quad
%         $^{2}$ TuSimple\\
%         $^{3}$ School of Artificial Intelligence, University of Chinese Academy of Sciences (UCAS)\\
%         $^{4}$ Centre for Artificial Intelligence and Robotics, HKISI\_CAS\\
%         {\tt\small
%  \{hejiawei2019, zhaoxiang.zhang\}@ia.ac.cn \{zehaohuang18, winsty\}@gmail.com}
%         }
% \maketitle

% \thispagestyle{empty}

% \newtheorem{theorem}{Theorem}[section]
\renewcommand\thesection{\Alph{section}} 
% \begin{document}
\pagestyle{empty}
\setcounter{section}{0}
    \setcounter{figure}{0}
    \setcounter{table}{0}
    \setcounter{equation}{0}
    \setcounter{footnote}{0}
\renewcommand{\thetable}{\Alph{table}}
\renewcommand{\thefigure}{\Alph{figure}}
\renewcommand{\theequation}{\Alph{equation}}
\renewcommand{\thealgocf}{\Alph{algorithm}}

% \thispagestyle{fancy}
%%%%%%%%% ABSTRACT
% \begin{abstract}
% In the Supplementary Material, we provide review of the original formulation of graph matching and some practically used QAPs, the connections between them.  
% \end{abstract}
%%%%%%%%% BODY TEXT
% \section{Appendix}
\appendix
% \section{Quadratic Assignment Problem and Graph Matching}
% \subsection{Lawler's QAP and Koopmans-Beckmann's QAP}
% \subsection{Detailed Derivations}

\section{Gradients of the Graph Matching Layer}
As described in Section {\color{red} 4.3} of our main paper, the gradients of the graph matching layer we need for backward can be derived from the KKT conditions with the help of the implicit function theorem. Here, we show the details of deriving the gradients. 
\par
For a quadratic programming (QP), the standard formulation is as
\begin{equation}
    \begin{aligned}
    & \minimize_x
    & & \frac{1}{2}x^\top Q(\theta) x + q(\theta)^\top x \\
    & \subjectto
    && G(\theta)x \leq h(\theta) \\
    &&& A(\theta)x=b(\theta).
    \end{aligned}
    \end{equation}
So the Lagrangian is given by 
\begin{equation}
    L(x,\nu,\lambda)=\frac{1}{2}x^\top Qx+\lambda^\top (Gx-h)+q^\top x+\nu^\top (Ax-b),
\end{equation}
where, $\nu$ and $\lambda$ are the dual variables.\\
The $(x^*,\lambda^*,\nu^*)$ are the optimal solution if and only if they satisfy the KKT conditions:
\begin{equation}
    \begin{split}
    \nabla_x L(x^*,\lambda^*,\nu^*) &= 0 \\
    Qx^* +q+A^\top \nu^*+G^\top \lambda^* &= 0 \\
    Ax^*-b &= 0 \\
    \diag (\lambda^*)(Gx^*-h) &= 0 \\
    Gx^* -h &\leq 0 \\
    \lambda^*&\geq 0.
    \end{split}
    \end{equation}
We define the function
\begin{equation}
    g(x,\lambda,\nu,\theta) = \begin{bmatrix}
    \nabla_x L({x},\lambda,\nu,\theta) \\
    \diag(\lambda)\lambda^\top (G(\theta)x-h(\theta)) \\
    A(\theta)x-b(\theta)
    \end{bmatrix},
    \end{equation}
and the optimal solution $x^*, \lambda^*, \nu^*$ satisfy the euqation $g(x^*, \lambda^*, \nu^*,\theta)=0$. \\
According to the implicit function theorem, as proven in \cite{barratt2018differentiability}, the gradients where the primal variable $x$ and the dual variables $\nu$ and $\lambda$ are the optimal solution, can be formulated as 
    \begin{equation}
        J_\theta x^* = -J_x g(x^*,\lambda^*,\nu^*,\theta)^{-1} J_\theta g(x^*,\lambda^*,\nu^*,\theta),
    \label{eq:jacobian}
    \end{equation}
where, $J_x g(x^*,\lambda^*,\nu^*,\theta)$ and $J_\theta g(x^*,\lambda^*,\nu^*,\theta)$ are the Jacobian matrices. Each element of them is the partial derivative of function $g$ with respect to variable $x$ and $\theta$, respectively.

\section{Pseudo-code of Our Algorithm}

To make our algorithm clear and easy to understand, we show the pseudo code of our GMTracker algorithm in Alg.~\ref{alg:gmtracker}.
The input of the algorithm is the detection set $\MCD^t=\{D_1^t, D_2^t,\cdots, D_{n_d}^t\}$ and tracklet set $\MCT^t=\{T_1^t, T_2^t,\cdots, T_{n_t}^t\}$, defined in Section {\color{red} 4.1} of our main paper. And the output is the new tracklet set $\MCT^{t+1}$ to be associated in the next frame. The motion gate $\kappa$ is 9.4877. The feature similarity threshold $\sigma$ is 0.6 in the videos taken by the moving camera, and 0.7 in the videos taken by the static camera. The max age $\delta$ is 100 frames.

\begin{table}
    \begin{center}
        \resizebox{\columnwidth}{!}{
      \begin{tabular}{c|cc|ccccc}
        \toprule
         &IDF1 & MOTA & MT & ML & FP & FN & ID Sw.\\
        \midrule
        Baseline &68.1 &  62.1  &  556 &  371  &  1923  &  124480  &  1135  \\
        Ours &71.5 & 62.3  &  555 &  375  & 1741  &  124298  & 1017  \\
        Oracle &77.2 &62.6 &545 &368 &1730 &124287 &14 \\
        \bottomrule
      \end{tabular}}
      \end{center}
    \caption{Comparison between the baseline, our GMTracker and the Oracle tracker on MOT17 \emph{val} set.}
  \label{tbl-oracle}
  \end{table}
\begin{figure}[t]
    \includegraphics[width=\columnwidth]{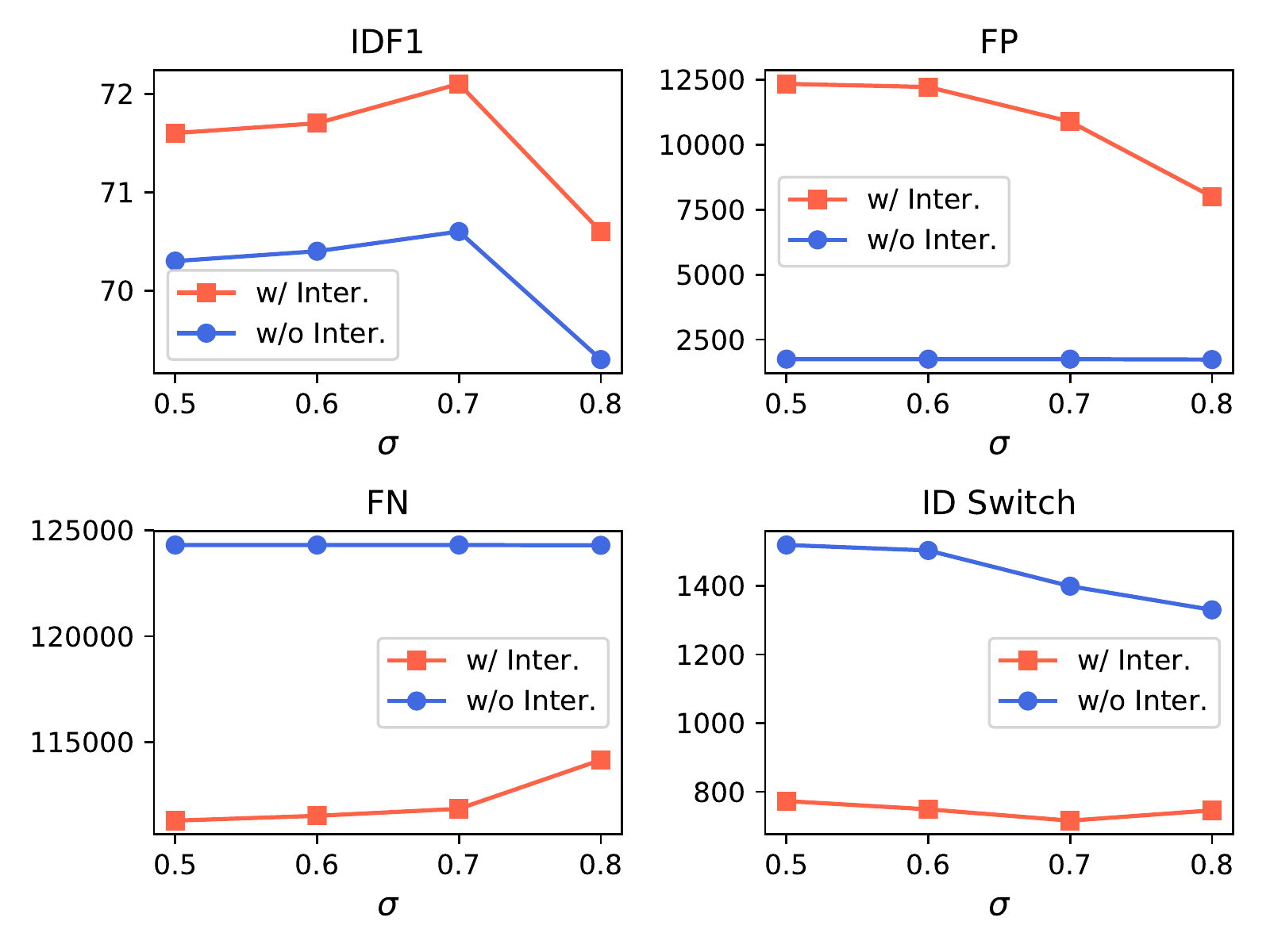}
    \caption{Results on IDF1, FP, FN and ID Switch metrics under different threshold $\sigma$ of the feature similarity to create a new tracklet.}
    \vspace{-0.2cm}
    \label{fig:thr}
    \vspace{-3pt}
\end{figure}
\section{Additional Experiments and Analyses}

% \subsection{Ablation of Camera Motion Compensation}

% \input{tables/w&wo-ecc.tex}

\subsection{Comparison with the Oracle Tracker}

To explore the upper bound of the association method, we compare our method with the ground truth association, called the Oracle tracker. The results on MOT17 \emph{val} set are shown in Table \ref{tbl-oracle}. There is a gap of 5.7 IDF1 and about 1000 ID Switches between our online GMTracker and the Oracle tracker. 
\par
Another observation is that on some metrics, which are extremely relevant to detection results, like MOTA, FP and FN, the gaps between the baseline, our method and the Oracle tracker are relatively small. That is why we mainly concern with the metrics reflecting the association results, such as IDF1 and ID Switch.

\subsection{Discussions}
\noindent{\bf Tracklet born strategies.}
In our GMTracker, the tracklet born strategies mostly follow DeepSORT, but we also make some improvements to make these strategies more suitable for our approach, as described in Section \textcolor{red}{4.4} in our main paper.
 Among the three criteria to create a new tracklet, we find that the threshold $\sigma$ is the most sensitive hyperparameter in our method. We conduct experiments with different $\sigma$, and its influence on IDF1, FP, FN and ID Switch is shown in Fig.~\ref{fig:thr}.
% \subsection{A Better Tracker Based on CenterTrack Detections}
% \subsection{Speed Analysis}
% \input{tables/centertrack.tex}

% \subsection{Detailed Results of Each Video on MOT Challenge Benchmark}
% The detailed results using our online and offline method based on Tracktor \cite{bergmann2019tracking} and CenterTrack \cite{zhou2020tracking} detections on MOT Challenge benchmark are shown in Table \ref{online}, \ref{offline} and Table \ref{cent}. The detailed results indicate that our method can achieve a better performance in the crowded scene than state-of-the-art trackers due to our robust matching algorithm.
\subsection{Detailed Performance}
As shown in Table~\ref{tab:detail-mot}, the results on more metrics, such as HOTA, AssA, DetA, LocA, MT, ML are provided for better comparison.

\renewcommand{\thealgocf}{A}
\begin{algorithm}[ht]
        \caption{GMTracker Algorithm}
        \label{alg:gmtracker}
      
        \KwIn{$\MCD^t,\MCT^t$}
        \KwOut{$\MCT^{t+1}$}
      
        \For{$D_i^t \in \MCD^t$}{
        $\mathbf{a}_D^{i,t} \leftarrow \mathtt{MLP_a}(\mathtt{ReID}(\mathbf{I}_i^t))$ \\
        $\mathbf{h}_i^{(0)} \leftarrow \mathbf{a}_D^{i,t}$
        }
        \For{$T_j^t \in \MCT^t$}{
            \For{$D_{(j)}^k \in T_j^t$}
            {$\mathbf{a}_D^{(j),k} \leftarrow \mathtt{MLP_a}(\mathtt{ReID}(\mathbf{I}_{(j)}^k))$}
            $\mathbf{a}_T^{j,t} \leftarrow \mathtt{mean}(\mathbf{a}_D^{(j),k})$ \\
            $\mathbf{h}_j^{(0)} \leftarrow \mathbf{a}_T^{j,t}$
            }
        \For{$l \leq l_{max}$}{
        \For{$D_i^t \in \MCD^t$}{
            $\Mm_{i}^{(l)} \leftarrow \MCA(\{w_{i,j}^{(l)}\Mh_j^{(l)} \mid j \in \MCG_T\})$ \\
            $\Mh_{i}^{(l + 1)}  \leftarrow \MCF(\Mh_i^{(l)}, \Mm_{i}^{(l)})$
        }
        \For{$T_j^t \in \MCT^t$}{
            $\Mm_{j}^{(l)}  \leftarrow \MCA(\{w_{i,j}^{(l)}\Mh_i^{(l)} \mid i \in \MCG_D\})$ \\
            $\Mh_{j}^{(l + 1)}  \leftarrow \MCF(\Mh_j^{(l)}, \Mm_{j}^{(l)})$
            }
        }
        \For{$D_i^t,D_{i'}^t \in \MCD^t$}{
            $\Mh_{i} \leftarrow \Mh_{i}^{(l + 1)}$, $\Mh_{i'} \leftarrow \Mh_{i'}^{(l + 1)}$\\
            $\mathbf{h}_{i,i'} \leftarrow l_2([\Mh_i, \Mh_{i'}])$
        }
        \For{$T_i^t,T_{i'}^t \in \MCT^t$}{
            $\Mh_{j} \leftarrow \Mh_{j}^{(l + 1)}$, $\Mh_{j'} \leftarrow \Mh_{j'}^{(l + 1)}$\\
            $\mathbf{h}_{j,j'} \leftarrow l_2([\Mh_j, \Mh_{j'}])$
        }
        \For{$D_i^t,T_j^t \in \MCD^t,\MCT^t$}{
            $\mathbf{M}_e^{u,v} \leftarrow \mathbf{h}_{i,i'}^\top\mathbf{h}_{j,j'}$\\
            $\mathbf{B}_{{i,j}}\leftarrow \mathbf{h}_{i}^\top\mathbf{h}_{j}$
        }
        $\mathtt{match} \leftarrow \mathtt{graph\_matching}(\mathbf{M}_e,\mathbf{B})$\\
        % $\mathtt{match} \leftarrow \mathtt{greedy}(\mathbf{X})$\\
        \For{$D_i^t,T_j^t \in \MCD^t,\MCT^t$}{
            \If{$\mathtt{IoU}(D_i^t,T_j^t)\leq 0\ \mathtt{or}\ \mathtt{d}(D_i^t,T_j^t) > \kappa\ \mathtt{or}\ \mathtt{cos}(D_i^t,T_j^t) < \sigma$}{$\mathtt{delete}(\mathtt{match}(i,j))$}

            }
            \For{$D_i^t,T_j^t \in \MCD^t_{\mathtt{unmatch}},\MCT^t_{\mathtt{unmatch}}$}{
                
            \If{$\mathtt{IoU}(D_i^t,T_j^t)\geq 0.3$}{
                $\mathtt{match}_\mathtt{add} \leftarrow \mathtt{Hungarian}(\mathtt{IoU}(D_i^t,T_j^t))$
            }}
            \For{$D_i^t,T_j^t \in \MCD^t,\MCT^t$}{
            \If{$\mathtt{match}(i,j)\ \mathtt{or}\ \mathtt{match}_\mathtt{add}(i,j)$}{
                $T_j^{t+1}\leftarrow T_j^t + \{D_i^t\}$\\
                $\mathtt{motion}(T_j^{t+1}).\mathtt{update()}$
            }
            \If{$D_i^t \in \MCD^t_{\mathtt{unmatch}}$}{$T_\mathtt{new}^{t+1} \leftarrow \{D_i^t\}$}
            
            \If{$T_j^t.\mathtt{last\_update}>\delta$}{$\mathtt{delete}(T_j^t)$}}
        \Return{$\MCT^{t+1}$}

    \end{algorithm}

% \input{tables/online.tex}
% \input{tables/offline.tex}
% \input{tables/cent_detail.tex}
% {\small
% \bibliographystyle{ieee_fullname}
% \bibliography{egbib}
% }
\clearpage
\begin{table*}[!t]
    \centering

    \tabcolsep=0.12cm
    \scalebox{0.9}{
        \begin{tabular}{l |c|c c c c c c c c c c c}
         \toprule
         Methods &Refined Det& IDF1 $\uparrow$ &HOTA$\uparrow$& MOTA $\uparrow$ &MT$\uparrow$ &ML$\downarrow$ & FP $\downarrow$ & FN $\downarrow$ & IDS $\downarrow$  &AssA$\uparrow$ &DetA$\uparrow$ &LocA$\uparrow$\\ [0.5ex] 
         \midrule
         \multicolumn{13}{c}{MOT17} \\
         \midrule
         GNMOT (O$^*$) \cite{Li_2020_WACV} &- &47.0&-&50.2 &19.3 & 32.7   &29316 &246200 &5273 &-&-&- \\
         FAMNet (O) \cite{chu2019famnet}&- &48.7&-&52.0 &19.1 &33.4  &14138 &253616 &3072 &-&-&-\\
         JBNOT (O$^*$) \cite{henschel2019multiple}&- &50.8&41.3&52.6 &19.7 &35.8 &31572 &232659 &3050 &39.8 &43.3 &80.2\\     
         Tracktor++ (O)  \cite{bergmann2019tracking}&Tracktor & 52.3&42.1& 53.5&19.5&36.6 & 12201 & 248047 & 2072 &41.7 &42.9 &80.9\\
         Tracktor++v2 (O)  \cite{bergmann2019tracking}&Tracktor& 55.1&44.8 & 56.3&21.1&35.3  & 8866 & 235449 & 1987&45.1&44.9 &81.8\\
         GNNMatch (O) \cite{papakis2020gcnnmatch}&Tracktor  &56.1 &45.4&57.0&23.3&34.6 &12283 	&228242 &1957&45.2 	&45.9  &81.5\\
         GSM\_Tracktor (O) \cite{liugsm}&Tracktor &57.8 &45.7&56.4&22.2&34.5&14379 &230174 &\textbf{1485} &47.0 &44.9 &80.9\\
         CTTrackPub (O) \cite{zhou2020tracking} &CenterTrack &59.6&48.2&61.5  &\textbf{26.4}  &\textbf{31.9} &14076 &200672 &2583 &47.8 &49.0 &81.7\\
         \bf{GMTracker(Ours)} (O) &Tracktor& 63.8 &49.1& 56.2  &21.0 &35.5 &\textbf{8719} & 236541 & 1778 &53.9 &44.9 &81.8 \\
         \bf{GMT\_CT(Ours)} (O) &CenterTrack& \textbf{66.9}&\textbf{52.0}& \textbf{61.5} &26.3 &32.1 & 14059 & \textbf{200655} & 2415 &\textbf{55.1} &\textbf{49.4} &\textbf{81.8} \\
         \midrule
         TPM \cite{peng2020tpm} &- &52.6 &41.5&54.2 &22.8&37.5  &13739 &242730 &1824&40.9 &42.5 &80.0\\
         eTC17 \cite{wang2019exploit}&-  &58.1 &44.9&51.9&23.1&35.5&36164 &232783 &2288&47.0 &43.3&79.4\\
         MPNTrack~\cite{braso2020learning} &Tracktor& 61.7 &49.0& 58.8&28.8&33.5 & 17413 &213594 &1185 &51.1&47.3&81.5\\
         Lif\_TsimInt~\cite{hornakova2020lifted}&Tracktor&65.2 &50.7&58.2 &28.6&33.6 &16850 &217944 &\textbf{1022} &54.9&47.1&81.5\\
         LifT~\cite{hornakova2020lifted}&Tracktor &65.6 &51.3&60.5 &27.0&33.6&14966 &206619 &1189&54.7&48.3&81.3\\
         \bf{GMT\_simInt (Ours)} &Tracktor&  65.9&51.1& 59.0 &29.0 &33.6&  20395 & 209553 & 1105 &55.1 &47.6 &81.2 \\
         \bf{GMT\_VIVE (Ours)} &Tracktor&65.9 &51.2 &60.2 &26.5 &33.2 &\textbf{13142} &209812 &1675 &55.1 &47.8 &81.3\\
         \bf{GMTCT\_simInt (Ours)}&CenterTrack &\textbf{68.7} &\textbf{54.0}&\textbf{65.0} &\textbf{29.4} &\textbf{31.6}&18213 &\textbf{177058} &2200 &\textbf{56.4} &\textbf{52.0} &\textbf{81.5}\\
         \midrule
         \multicolumn{13}{c}{MOT16} \\
         \midrule
         Tracktor++v2 (O)  \cite{bergmann2019tracking} &Tracktor&54.9 &44.6&56.2&20.7&35.8 &2394 	&76844 	&617 &44.6&44.8&82.0\\
         GNNMatch (O) \cite{papakis2020gcnnmatch} &Tracktor&55.9  &44.6&56.9&22.3&35.3&3235 	&74784 	&564 &43.7&45.8&81.7\\
         GSM\_Tracktor (O)\cite{liugsm} &Tracktor &58.2&45.9 &57.0  &22.0&34.5 &4332 &73573 &\textbf{475} &46.7&45.4&81.1\\
         \bf{GMTracker(Ours)} (O)  &Tracktor&  63.9&48.9& 55.9 &20.3 &36.6 & \textbf{2371}& 77545 & 531  &53.7 &44.6 &\textbf{82.1}\\
         \bf{GMT\_CT (Ours)} (O)&CenterTrack &  \textbf{68.6} &\textbf{53.1}&\textbf{ 62.6} &\textbf{26.7} &\textbf{31.0} &  5104 &  \textbf{62377} & 787 &\textbf{56.3}&\textbf{50.4}  &81.8 \\
         \midrule
         TPM \cite{peng2020tpm} &- &47.9 &36.7 &51.3&18.7&40.8	&\textbf{2701} &85504 &569&34.6&39.3&79.1\\ 
         eTC \cite{wang2019exploit} &-  &56.1&42.0&49.2 &17.3&40.3	&8400 	&83702 &606 &44.5&39.9&78.8\\
         MPNTrack~\cite{braso2020learning}&Tracktor  & 61.7&48.9& 58.6  &27.3&34.0 & 4949 &70252 &354&51.1&47.1&81.7\\
         Lif\_TsimInt~\cite{hornakova2020lifted} &Tracktor &64.1 &49.6&57.5 &25.4&34.7&4249 &72868 &\textbf{335}&53.3&46.5&\textbf{81.9}\\
         LifT~\cite{hornakova2020lifted}&Tracktor&64.7 &50.8&61.3 &27.0&34.0 &4844 &65401 &389&53.1&48.9&81.4\\
         \bf{GMT\_simInt (Ours)} &Tracktor&  66.2 &51.2& 59.1  &27.5 &34.4 &  6021 &  68226 & 341 &55.1 &47.7 &81.5 \\
         \bf{GMT\_VIVE (Ours)}&Tracktor&66.6 &51.6&61.1 &26.7 &33.3&3891 &66550 &503 &55.3 &48.5 &81.5\\
         \bf{GMTCT\_simInt (Ours)}&CenterTrack  &\textbf{70.6}&\textbf{55.2} &\textbf{66.2}&\textbf{29.6} &\textbf{30.4} &6355 &\textbf{54560} &701 &\textbf{57.8} &\textbf{53.1} &81.5\\
         \bottomrule
        \end{tabular}}
    
    \caption{Detailed comparison with state-of-the-art methods on MOT16 and MOT17 \emph{test} set. (O) denotes online methods. (O$^*$) denotes near-online methods.}
    \label{tab:detail-mot}
    \end{table*}
% \end{document}

\clearpage
{\small
\bibliographystyle{ieee_fullname}
\bibliography{egbib}
}
\clearpage
\end{document}